\documentclass[a4paper,fleqn]{cas-dc}

\usepackage[authoryear]{natbib}
\usepackage{bbding}
\usepackage{subfigure}

\def\tsc#1{\csdef{#1}{\textsc{\lowercase{#1}}\xspace}}
\tsc{WGM}
\tsc{QE}

\begin{document}
\let\WriteBookmarks\relax
\def\floatpagepagefraction{1}
\def\textpagefraction{.001}

\shorttitle{BasicTAD: an Astounding RGB-Only Baseline for Temporal Action Detection}    

\shortauthors{Min Yang, Guo Chen, Yin-Dong Zheng, Tong Lu and Limin Wang}  

\title [mode = title]{BasicTAD: an Astounding RGB-Only Baseline for Temporal Action Detection}  



%

\author[1]{Min Yang}
\fnmark[1]
\affiliation[1]{organization={The State Key Laboratory for Novel Software Technology, Nanjing University},
            city={Nanjing},
            postcode={210023}, 
            country={China}}

\author[1]{Guo Chen}
\fnmark[1]

%
\author[1]{Yin-Dong Zheng}

\author[1]{Tong Lu}

\author[1]{Limin Wang \corref{cor1}}

\cortext[cor1]{Corresponding author (lmwang@nju.edu.cn).}

\fntext[1]{M. Yang and G. Chen equally contribute to this work.}


\begin{abstract}
Temporal action detection (TAD) is extensively studied in the video understanding community by generally following the object detection pipeline in images. However, complex designs are not uncommon in TAD, such as two-stream feature extraction, multi-stage training, complex temporal modeling, and global context fusion. In this paper, we do not aim to introduce any novel technique for TAD. Instead, we study a simple, straightforward, yet must-known baseline given the current status of complex design and low detection efficiency in TAD. In our simple baseline (BasicTAD), we decompose the TAD pipeline into several essential components: data sampling, backbone design, neck construction, and detection head. We extensively investigate the existing techniques in each component for this baseline and, more importantly, perform end-to-end training over the entire pipeline thanks to the simplicity of design. As a result,  this simple BasicTAD yields an astounding and real-time RGB-Only baseline very close to the state-of-the-art methods with two-stream inputs.
In addition, we further improve the BasicTAD by preserving more temporal and spatial information in network representation (termed as PlusTAD).
Empirical results demonstrate that our PlusTAD is very efficient and significantly outperforms the previous methods on the datasets of THUMOS14 and FineAction. Meanwhile, we also perform in-depth visualization and error analysis on our proposed method and try to provide more insights into the TAD problem. Our approach can serve as a strong baseline for future TAD research. The code and model are released at \url{https://github.com/MCG-NJU/BasicTAD}.
\end{abstract}


\begin{keywords}
 \sep Temporal Action Detection \sep End-to-end training \sep Baseline
\end{keywords}

\maketitle

\section{Introduction}\label{intro}
Video understanding is a fundamental and challenging problem in computer vision research. Temporal action detection (TAD)~\citep{THUMOS14,anet,fineaction}, which aims to localize the temporal interval of each action instance in an untrimmed video and recognize its action class, is particularly crucial for long-term video understanding. Due to the high complexity of video and the lack of a clear definition of action boundaries~\citep{moltisanti2017trespassingboundary}, past efforts on TAD often employ a relatively sophisticated paradigm to solve this problem. For example, they typically use two-stream inputs for feature extraction in advance~\citep{tcanet,a2net,rtd,pbrnet,afsd,sparse-rcnn-tad}, a multi-stage training strategy for different components~\citep{bsn,rtd}, temporal reasoning with complex models like graph neural networks~\citep{g-tad,DCAN,pgcn}, and yielding the detection results with the global classification scores~\citep{bmn,rtd,bsn} (e.g., UntrimmedNet~\citep{untrimmednet}). 
Such complex designs need additional extraction of optical flow, action detectors trained on temporal features rather than raw video frames, complex components for multi-stage training and additional classification results. Unlike object detectors in images, these complex designs hinder the existing TAD methods from being a simple and neat detection framework that could be easily trained in an end-to-end manner and efficiently deployed in real-world applications. 

Given the current status of complex designs and low detection efficiency in the TAD method, there is a need to step back and reconsider the basic principle of designing an efficient and practical TAD framework. In order to cater to the need for scalability, ease of use, and deployment into real applications for real needs, {\bf simplicity} is the most important requirement for building an efficient and practical TAD method. Inspired by the great success of object detectors~\citep{focalloss,fcos}, we argue that an optimal action detector should strictly follow a similar {\em modular design}, where each component has its function, and different components can be easily integrated seamlessly. In addition, {\bf efficient training} is another important property that needs to be considered. Multi-stage training often brings extra computational and storage costs while not being conducive to unleashing the power of deep learning. We argue that an ideal action detector should be end-to-end trainable where the entire pipeline trains and infers directly against raw video frames. Finally, being {\bf free of pre-processing} is another desired property for action detector design, particularly important for the deployment in real applications. We argue that optical flow extraction is the bottleneck for many TAD methods as it often requires a high computational cost to calculate these inputs explicitly. An RGB-only TAD approach should be more favorable and practical for video understanding applications.

Therefore, it is urgent to re-design a simple baseline for Temporal Anomaly Detection (TAD) to meet the abovementioned requirements. In this paper, we do not aim to introduce any new techniques for building a TAD framework. Rather, we investigate a simpler yet crucial baseline for the TAD task. Unlike the complex designs used in existing TAD frameworks, we carefully design a modular temporal detection framework that enables us to conduct in-depth studies on different components to determine the optimal settings, including data augmentation, backbone, neck, and detection head. While the design for standard object detectors has been highly mature and robust, better practice for building a temporal detector in videos is still needed. For each module design, we choose the simplest and most basic options and extensively investigate them to discover the optimal configurations. Additionally, we explore different data sampling and augmentation techniques during training and testing to develop an effective TAD method, which has been largely overlooked in previous methods. Based on our modular design and extensive empirical investigation, we establish an RGB-only baseline for TAD called BasicTAD. Our BasicTAD achieves competitive performance compared to the state-of-the-art methods with two-stream inputs. This performance indicates that the basic design of TAD requires reconsideration, and the search for the optimal basic design is foundational work that must be taken into account.

Encouraged by the outstanding performance of BasicTAD, we strictly follow its basic principle of designing an ideal TAD method and then further improve it minimally but meaningfully. Our core idea is to preserve complete temporal information in our backbone and spatial information in our neck under enhanced data augmentation. The implementation is quite simple to achieve these objectives by removing the temporal downsampling operations in the backbone and exchanging the spatial and temporal pooling operations in the neck. Such a simple design allows us to further explore larger temporal and spatial inputs for better detection results. These small changes would significantly improve the performance of BasicTAD, increasing the performance from 50.5\% to 59.6\% mAP on the THUMOS14 dataset. The resulted TAD method is denoted by \textbf{PlusTAD}, and we ascribe its good performance to our core idea of keeping rich information and careful implementation. We believe our work is timely and will draw the whole TAD community to reconsider the design of TAD methods, given the current status of complex design and low efficiency. The basic comparisons of our BasicTAD and PlusTAD with previous end-to-end TAD methods are summarized in Table~\ref{table:methods}. In summary, our contributions are threefold:
\begin{itemize}
    \item We reconsider the TAD pipeline and present a simple modular detection framework. For each component in our modular framework, we perform extensive studies on the existing basic choices. Through extensive empirical studies, we have developed good practice to build an astounding RGB-Only baseline method for TAD, called BasicTAD.
    \item Encouraged by the outstanding performance of BasicTAD, we further improve it with minimal changes to fully unleash the power of deep networks. Our core idea is to preserve richer information during our backbone feature extraction and neck compression. This idea could be easily implemented with high efficiency and the resulting PlusTAD significantly improve the TAD performance on the THUMOS14 dataset.
    \item The extensive experiments on THUMOS14 and FineAction demonstrate that PlusTAD outperforms previous state-of-the-art methods by a large margin. In particular, we obtain an average mAP of 59.6\% on the THUMOS14 only with RGB input. In addition, we perform in-depth ablation studies and error analysis to provide more insights for the future TAD pipeline design. 
\end{itemize}

\begin{table*}[t]
\centering
\caption{
\textbf{Comparison with previous end-to-end TAD methods only with RGB input on THUMOS14~\citep{THUMOS14} dataset.}
We categorize components and settings based on their order in the whole pipeline:
i) Data Stream: modal, resolution in temporal and spatial;
ii) Network: The backbone with $\beta$ times temporal downsampling ($\times\beta$) for feature extraction, the Neck for feature aggregation, and the Head for detecting temporal action segments. The temporal downsample module (TDM), and temporal feature pyramid network (TFPN) in Neck are two different methods for generating multi-scale features in the subsequent discussion. SP-NECK means our improved neck module.
iii) Performance: The speed metric is FPS. mAP represents the average mAP from mAP@0.3 to mAP@0.7. 
}
\label{table:methods}
\small
\setlength\tabcolsep{2.1mm}
\resizebox{1.0\textwidth}{!}{
\begin{tabular}{c|c|c|c|c|c}
\hline
\toprule
  \multirow{2}{*}{Method}&R-C3D&AFSD  &DaoTAD & \textbf{BasicTAD} & \textbf{PlusTAD} \\
    &~\citep{r-c3d}&~\citep{afsd}  &~\citep{rgb_enough} & (Ours) & (Ours) \\
 \toprule
  \multicolumn{6}{c}{Data Stream} \\
  \hline
 Modality & RGB & RGB & RGB & RGB & RGB \\
  \hline
 & 768 & 256 & 768 & 768 & 96 \\
Resolution & 25 FPS & 10 FPS & 25 FPS & 24 FPS & 3 FPS \\
  & 112$\times$112 & 112$\times$112 & 128$\times$128 & 128$\times$128 & short-128 \\
  \toprule
  \multicolumn{6}{c}{Network} \\
  \hline
Backbone & C3D($\times$8) & I3D($\times$8) & R50-I3D($\times$8) & SlowOnly($\times$8) & \textbf{TP-SlowOnly}($\times$1) \\
Neck & TDM & TFPN & TDM+TFPN & TDM+TFPN/TDM & \textbf{SP-NECK} \\
Head & R-C3D & AFSD  & RetinaNet & \textbf{Anchor-based}$/$\textbf{Anchor-free} & \textbf{Anchor-based}$/$\textbf{Anchor-free}\\ 
  \toprule
\multicolumn{6}{c}{Performance} \\
  \hline
mAP & $<$ 36.4 & 43.6  & 50.0 & 50.2$/$50.5 & 54.9$/$54.5 \\
Speed & 1030&  4057  & 6668 & 5454$/$2702 & 17454$/$8377\\   \hline
\end{tabular}
}
\end{table*}

\section{Related Work}
\label{sec:relatedwork}
\subsection{Action Recognition}  
Action recognition is an important task in video understanding.
Current deep learning-based action recognition methods can be mainly divided into two types. The first one is the CNN-based method, which includes the specific video architectures of 2D CNN, 3D CNN, and (2+1)D CNN. 2D CNN methods~\citep{two-stream,tsn,tdn} take RGB frames and optical flow as input to capture appearance and motion information, respectively. 
3D CNN methods~\citep{c3d,r3d,i3d,artnet,r50-i3d-non-local,slowfast} capture spatiotemporal information between frames by performing 3D convolution on stacked video frames. (2+1)D CNN methods~\citep{p3d,s3d,r2+1d,tsm,teinet,tea,tam} model spatiotemporal features by decoupling 3D convolution into 2D convolution and 1D convolutions or efficient temporal modules for reducing the computational complexity of the network. 

The second type of neural network is the Transformer~\citep{transformer} architecture, which successfully uses a global self-attention mechanism to address the limitation of CNN in an insufficient receptive field. The great success of image transformers~\citep{vit,deit,swin-transformer} has led to the investigation of video transformers~\citep{timesformer,video-swin-transformer,vtn,vivit,videomae,VideoMAEv2} for action recognition in videos. 
However, compared with CNN-based methods, the quadratic complexity of self-attention operations of Transformer-based architectures has led to high training costs and memory consumption. These research efforts on backbone design are orthogonal to our study on the TAD task. Any video backbone could be compatible with our BasicTAD and PlusTAD designs. In this paper, we explore the commonly-used action recognition backbone~\citep{c3d,i3d,r50-i3d-non-local,slowfast} and mainly choose SlowOnly~\citep{slowfast} to generate the spatiotemporal features due to its good trade-off between accuracy and efficiency.

\subsection{One-stage Temporal Action Detection}  
One-stage TAD methods aim to detect the boundaries and categories of action segments in a single shot. The existing one-stage TAD methods can be divided into anchor-based ones \citep{rgb_enough,ssad,end2end,gtan,a2net} and anchor-free ones~\citep{a2net,actionformer}. Most existing methods are anchor-based. For example, \cite{ssad} presented the first one-stage TAD method using convolutional networks. \citep{gtan} proposed to use Gaussian kernels to optimize the scale of each anchor dynamically. Meanwhile, ~\cite{rgb_enough} explored the pipeline of RetinaNet~\citep{retina} in the TAD task with an RGB-only stream. Some works explore the application of anchor-free methods. For instance, ~\cite{a2net} explored the combination of anchor-based and anchor-free methods. ~\cite{actionformer} proposed to use a local transformer encoder as a neck to enhance the video features for TAD.

Our work shares the advantage of simplicity with these one-stage TAD methods by focusing on designing an end-to-end TAD baseline. However, our BasicTAD presents a modular design for easy systematic study over the entire TAD pipeline. Based on this pipeline, we perform a more extensive study on the entire pipeline's components and figure out a simpler yet must-known TAD baseline (BasicTAD). In addition, we further empower our BasicTAD by keeping complete temporal and spatial information to yield our final TAD method of PlusTAD.

\subsection{Multi-stage Temporal Action Detection}
Multi-stage TAD methods often involve multiple stages to generate and refine action detection results. These methods might focus on different aspects to obtain better detection results. Most of them~\citep{bsn,bmn,bsn++,g-tad,bcgnn,DCAN} focus on improving the quality of generated proposals, while a few others~\citep{TADTR,ssn} try to improve the quality of classification results. ~\citep{bsn,bmn,bsn++,g-tad,bcgnn, DCAN,turn,rapnet,talnet,r-c3d} generated candidate action proposals at first, which is called temporal action proposal generation, and then further classified them into action categories possibly with the global classification results (e.g., UntrimmedNet~\citep{untrimmednet}). For proposal generation, ~\cite{bsn,bmn,bsn++,g-tad,bcgnn, DCAN} were boundary-based methods that predict each frame's start and end confidence and then match start and end frames to generate the proposals with confidence evaluation.  ~\citep{turn,rapnet,talnet,r-c3d} generated proposals based on pre-defined sliding window anchors and trained a classifier to filter anchors. ~\citep{afsd} designed a saliency-based refinement module to refine the detection results generated by a one-stage detector. ~\cite{sparse-rcnn-tad, TADTR,rtd} adopted different query-based dynamic networks along with multi-stage refinement modules to generate a direct sparse action proposal, effectively removing the post-processing steps of NMS. To improve classification results, ~\cite{ssn} proposed regressing another completeness score to complement the classification score. ~\cite{TADTR} designed a post-processing technique to refine the confidence score based on actionness regression.

Unlike these multi-stage TAD methods, our BasicTAD aims to provide a one-stage and end-to-end TAD baseline method without requiring multi-stage processing. This simpler design would allow us to focus on an extensive investigation of basic yet overlooked designs of the TAD pipeline. As a result, we obtain a much simpler yet must-known TAD method, which obtains significant improvement over these complex multi-stage TAD methods on the THUMOS14 benchmark.

\subsection{Training Strategies for Temporal Action Detection}
There are various training strategies for optimizing TAD frameworks. The entire training process of TAD is usually split into multiple independent steps to reduce the optimization difficulty.
Mainstream training strategies can be divided into two types. 
The first type consists of two steps~\citep{bsn,bmn,bsn++,rapnet, TADTR,rtd}. 
First, the backbone networks are pre-trained on the TAD or action recognition datasets and used to perform feature extraction in a sliding window manner. 
Then, a separate TAD head network is trained without fine-tuning over the backbone networks to generate action segment proposals or directly predict the action segment boundaries and categories.
This multi-step training paradigm would fail to unleash the potential of end-to-end representation learning on TAD.

The second type is to train the whole TAD pipeline (i.e., backbone and detection head) from RGB frames or both RGB frames and optical flow in an end-to-end manner.
Recently, several works~\citep{ssad,sparse-rcnn-tad,afsd,rgb_enough,e2e-TADTR}, adopted this training strategy. Among them, the concurrent work~\citep{e2e-TADTR} compared the difference between head-only training and end-to-end learning and explored different backbones and detection heads. It also tried to balance detection results and computing overhead. However, it ignored the detailed investigation of some basic components in the entire TAD pipeline, such as the data augmentation and the neck design. Meanwhile, our result is better than its performance on the THUMOS14 dataset thanks to our simpler design and more detailed empirical study.

Our work shares the same advantage of end-to-end training with these methods. Compared with these works, our work conducts more detailed studies on the overall pipeline of end-to-end TAD and includes in-depth investigation for each component. Our solution is simpler yet more effective, achieving much better performance than these methods. BasicTAD could be easily extended to PlusTAD and has a fast inference speed, meeting the requirement of real-time TAD. We hope our extensive study could encourage future research to focus on designing an end-to-end TAD pipeline.

\section{Methodology}
\label{sec:methodology}
In this section, we reconsider the TAD pipeline design by focusing on simplicity, efficient training, and eliminating complex pre-processing steps.
We present a modular TAD framework consisting of four key components: data sampling (augmentation), backbone design, neck construction, and detection head. These components can be seamlessly integrated to yield a simple and efficient TAD framework. 
For each component, we explore the basic options to determine the optimal configuration, resulting in the {\bf BasicTAD} framework. Additionally, we make minimal modifications to the BasicTAD framework by proposing three new design principles and creating an improved version of {\bf PlusTAD}. Thanks to the straightforward design of both BasicTAD and PlusTAD, both frameworks enjoy end-to-end training and fully unleash the power of representation learning for the TAD task.

\begin{figure*}[!t]
  \includegraphics[width=1\textwidth]{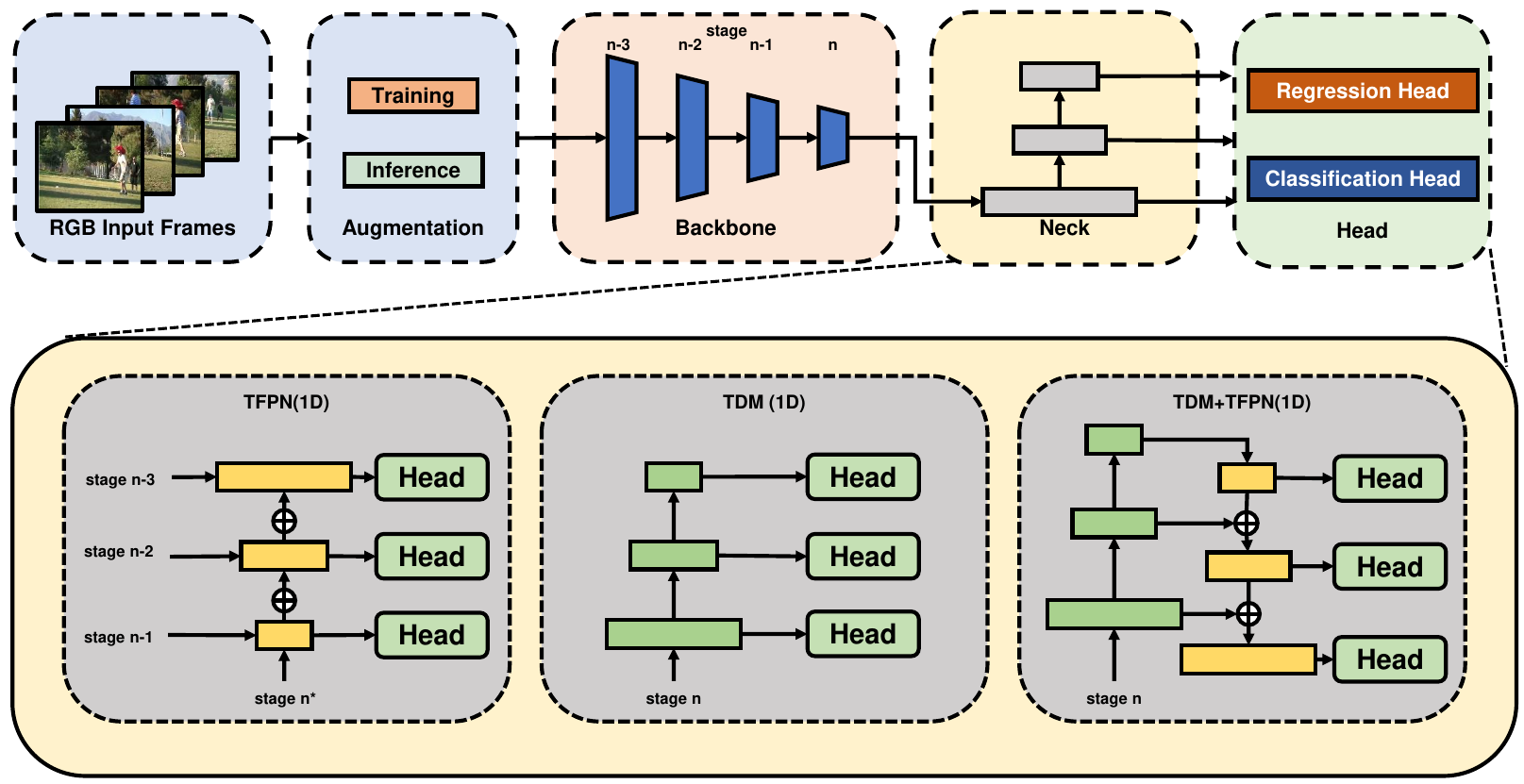}
  \caption{
\textbf{BasicTAD Pipeline.}
Our BasicTAD exhibits a modular design framework for the TAD task, composed of data sampling (augmentation), backbone, neck, and detection head.
(a) The backbone is a spatial-temporal network to extract temporal features.
(b) The neck module of BasicTAD offers three different ways of construction to leverage the extracted features. In the neck module, $n$ means the number of stages in the backbone, and stage $n$ is the last stage of the backbone. Stage $n*$ contains extra pooling operations on stage $n$ to construct a temporal feature with low temporal resolution. Particularly, $\bigoplus$ represents adding two features. As the temporal dimension of features varies across different scales, we interpolate the high-level features along the temporal dimension to align them with the low-level features before addition. (c) We adopt a typical one-stage network as the detection head, implemented by anchor-based and anchor-free methods. 
}
  \label{fig:arch-basic-TAD}
\end{figure*}

\subsection{BasicTAD}
 \label{basictad}
\label{sec:basictad}
To establish a straightforward and universal pipeline for facilitating the analysis and development of TAD methods, we break down the TAD pipeline into four fundamental components according to their functions: data augmentation, backbone design, neck construction, and one-stage detection head design. In general, these components are in analogy with the design in the common object detectors~\citep{fcos, retina}. An overview of the modular TAD framework is illustrated in Fig~\ref{fig:arch-basic-TAD}, and we will delve into each component in detail in the following subsections.

\subsubsection{Data Sampling and Augmentation}
\label{sec:dataaugmentation}
As the TAD task involves temporal localization and prediction, the detection precision greatly relies on the robustness of the extracted feature sequence. The temporal sampling method is a crucial factor that affects feature extraction. The most basic sampling method is dense sampling with a fixed sampling rate. However, due to the limited memory of GPUs, it is not feasible to load an untrimmed video in its entirety. Therefore, we employ a sliding window strategy to divide the untrimmed video into overlapping clips. Within each video clip, we utilize the simple dense sampling method to downsample the original frame sequence with a fixed frame rate per second (FPS). This hyper-parameter FPS must be tuned to balance the detection accuracy and memory consumption afforded by the common GPU devices. We carefully tune these basic sampling options in our BasicTAD pipeline to demonstrate their influence on the final detection performance, which has been largely ignored in the existing TAD methods.

Another vital factor in feature extraction is data augmentation. Drawing on the success of multiple image-level data augmentation techniques in action recognition, we incorporate a similar data augmentation strategy in the first step of BasicTAD. Random cropping and horizontal flipping have been commonly utilized in existing end-to-end work, and we use the two methods by default. Furthermore, since video data may contain objects with various poses and scenes with different brightness levels, we adopt more image-based data augmentation techniques in our BasicTAD, including photo distortion and random rotation. In subsequent experiments, we employ all four data augmentation methods mentioned above. It is worth noting that although these augmentation techniques are commonplace in images, their effects on TAD tasks have yet to be explored in previous works.

\subsubsection{Backbone}
\label{sec:backbone}
The second component in BasicTAD is the backbone which is responsible for extracting spatiotemporal features from videos using a multi-layer network. However, the temporal modeling of early backbones only learns a simple consensus of RGB scene changes, making it suffering to capture temporal dynamics. 
Therefore, optical flow is naturally introduced in the previous TAD methods to explicitly supplement the short-form motion information. Nowadays, the latest designs of 3D backbones like SlowOnly can implicitly model such temporal dynamics of RGB input by learning to attain good performance on Something-something V2~\citep{sth-sth}. Hence, we believe that RGB input is enough, and we adopt an off-the-shelf 3D action recognition backbone to encode temporal dynamics for RGB-only input.

To balance the computing overhead and detection precision, we choose widely used SlowOnly~\citep{slowfast} as the backbone of BasicTAD by default.
For the early TAD work, $8\times$ spatial-temporal downsampling operation is applied in the backbones such as C3D~\citep{c3d} and I3D~\citep{i3d} to reduce the temporal resolution for saving computational overhead and increasing the receptive field to capture actions at larger scales. 
Since SlowOnly does not contain $8\times$ temporal downsampling, we insert three $2\times$ temporal downsampling operations to align the settings and fairly compare with these methods. We will perform ablation studies to explore the impact of the temporal downsampling operation.
We also perform comparative studies over the different choices of video backbones on the final TAD performance.

\subsubsection{Neck}
\label{sec:neck}
The neck module, located after the backbone and before the detection head, plays a critical role in the TAD task by facilitating the alignment between the video features and downstream detection tasks. 
Its objective is to construct multi-resolution representations that can flexibly handle the vast variety of temporal durations of action instances.
In this section, we introduce three neck modules and consider them candidates for the neck in BasicTAD.

After obtaining the extracted spatiotemporal features $F_{3d} \in R^{C\times T \times H \times W}$, we squeeze the H and W dimensions using spatial average pooling to get the temporal feature $F_{1d} \in R^{C\times T}$. We adopt multi-scale temporal features for our subsequent one-stage detection heads, helping the one-stage methods (anchor-based or anchor-free) better detect action segments of many scales. We use two basic feature pyramid networks, \emph{i.e.}, Temporal Feature Pyramid Network (TFPN) and Temporal Down Network (TDM) to build necks for creating or enhancing multi-scale temporal features from the backbone. TFPN enhances multi-scale temporal features by performing up-sampling and lateral addition on given multi-scale temporal features. Formally, the operation of each layer in TFPN can be represented as follows:
\begin{equation}
    F_{i}^{'}= \text{Conv}(F_{i}+\text{Upsample}(F_{i+1})),
\end{equation}
where $F_i$ represents the features in the $i$-th layer of the temporal feature pyramids, and $F_{i}^{'}$ denotes the new features that aggregate both high-level features $F_{i+1}$ and current scale features $F_{i}$. TDM utilizes multiple down-sampling operations to create multi-scale temporal features from a single-scale feature. Given a single-scale temporal feature sequence as the first-layer feature of feature pyramids, each layer of TDM can be expressed as follows:
\begin{equation}
    F_{i+1}=\text{Downsample}(F_{i})
\end{equation}
where $\text{Downsample}$ represents a max-pool or convolution layer, aggregating the temporal semantic information of the features and halving the temporal dimension. Based on both basic networks, as shown in Fig~\ref{fig:arch-basic-TAD}, we construct three necks: Lateral-TFPN, Post-TDM, and Post-TDM-TFPN.

\textbf{Lateral-TFPN.}
One of the most intuitive approaches to building a TFPN is by leveraging multi-scale spatiotemporal features extracted from different stages of the backbone. This approach facilitates the integration of high-level semantic information with low-level features, much like what is accomplished by FPN in object detection.
Although shallow layers offer a greater temporal resolution, the spatial features at each temporal location may need to capture more rich temporal semantic information, which could hinder the effectiveness of the TAD task.

\textbf{Post-TDM.} 
TDM generates multi-scale temporal features on pre-extracted temporal features. While the last layer of backbone features provides rich semantic information at each temporal location, increasing the number of feature layers may prevent the learned high-level semantics from being fed back to low-level features. TDM, on the other hand, utilizes features from the backbone directly and is, therefore, easier to optimize compared to TFPN.

\textbf{Post-TDM-TFPN.} 
Constructing TDM and TFPN after the backbone can simultaneously address the abovementioned problems. TFPN integrates high-level semantic information into low-level features, which could improve the network's capacity to aggregate temporal context while ensuring adequate spatial semantics. 
However, the introduction of additional learnable parameters could impede network optimization.  

In the subsequent experiments, we study and analyze the different choices of the three necks and provide our final combination in BasicTAD.

\subsubsection{Head}
\label{sec:label}
The detection head is the final component in our modular TAD framework. It is responsible for completing the detection task by generating the temporal interval of the action instance and its corresponding label. Typically, this component includes sub-networks designed for classification and regression tasks, respectively. Moreover, the specific sample assignment is critical for training these detection heads effectively. 
Both aspects complement each other to produce the best possible results in the final detection performance.

For simplicity and end-to-end training, we adopt anchor-based and anchor-free mechanisms as the basic detection head of our BasicTAD, by following the basic design principle in object detection~\citep{faster-rcnn,retina,fcos}.
Since these two detection methods have advantages and disadvantages, there is no clear conclusion on which is better. 
We briefly introduce two detection methods in our TAD pipeline and provide specific implementation details. Both methods share the same sub-networks composed of four temporal convolutional layers followed by a normalization and activation layer. Both methods share two classification and regression sub-networks composed of four temporal convolutional layers followed by a normalization and activation layer.
The regression and classification branches for both methods can be formulated as follows:
\begin{equation}
    F_{i}=\text{Conv}(F_{i-1}),
\end{equation}
where $i \in [1,4]$ is the output features of the $i$-th convolution layer and $F_{0}$ represents input features of head. 
In the last layer, the output of the classification branch is represented as $F_{4}^{cls} \in R^{N_a \times N_c \times T}$, where $N_a$ represents the number of anchors ($N_a = 1$ for anchor-free method) and $N_c$ represents the number of categories. The output of the regression branch denotes $F_{4}^{reg} \in R^{N_a \times 2 \times T}$, where $2$ represents the variables related to boundaries.
Since the sub-network design of both methods is identical, we will primarily focus on introducing the two methods from the perspective of the sample assignment mechanism.

\textbf{Anchor-based Method.}
Anchor-based methods generate temporal proposals by assigning dense and multi-scale intervals with pre-defined lengths to uniformly distributed temporal locations in the input video.
We use translation-invariant anchors, which have an increasing temporal size from the bottom to the top of the feature pyramid network.
At each level, we add anchors of 5 sizes 
$\left\{2^{0},2^{1/5},2^{2/5},2^{3/5},2^{4/5} \right\}$ of the original set of default anchors for dense scale coverage.
Anchors are assigned to ground-truth action segments using the temporal Intersection-over-Union (tIoU) threshold of 0.6 and a background with a tIoU lower than 0.4.
Other anchors overlapping [0.4, 0.6) will be eliminated during training. We obtain the predicted action boundaries by optimizing the relative offset between anchors and ground truths. Dense anchor matching is required at each temporal position of the feature for anchor-based methods, which need rich semantic context information.

\textbf{Anchor-free Method.}
Anchor-free methods directly regress the offsets to action boundaries at each temporal location and then use these offsets to generate temporal proposals. We first compute the regression offsets for each location on all feature levels.
Any location which falls into any ground-truth box will be set as a positive sample. The others are negative samples.
Each level is responsible for a range of motion detection.
If an action proposal at one temporal location is beyond this range, this location will be ignored. We define $m_i$ as the maximum range border that the feature level $i$ needs to regress. In this work, $m_2, m_3, m_4, m_5, m_6, m_7$ are set as $-1, 5, 10, 20, 40$ and $\infty$, respectively. Even with multi-level prediction, if a location is still assigned to more than one ground-truth box, we choose the ground-truth box with minimal area as its target. Subsequently, we employ $e^{s_ix}$ with a trainable scalar $s_i$ to automatically adjust the base of the exponential function for feature level $i$. This approach enables us to obtain the predicted actions' boundaries accurately. Compared with anchor-based methods, anchor-free methods have a more flexible matching mechanism. 

\textbf{Post Processing.}
We also apply post-processing to suppress redundant predictions to yield the final detection results for both anchor-free and anchor-based methods. Specifically, we choose two suppression algorithms: Non-Maximum Suppression (NMS)~\citep{nms} and Non-Maximum Weighting (NMW)~\citep{nmw} for both anchor-based and anchor-free methods. NMS is a crucial technique that ensures the algorithm produces only one detection per object. It selects the proposal with the highest confidence in each iteration, chooses all remaining proposals with a high overlap rate with the current proposal, and then suppresses them. After that, it saves the currently selected proposal and proceeds to the next iteration, excluding the selected proposal. This procedure continues until all proposals have been processed. NMW is an improved version of NMS since proposals with the highest confidence scores may not be accurately positioned, while other well-located proposals may exist. NMW uses the confidence score and Intersection over Union (IoU) to calculate a weighted average of all proposal coordinates of the same type. We will conduct ablation studies on them in the later section.

The above detection methods and post-processing mechanisms are our optional basic components in head.
Validation and more in-depth studies on these components will be shown in ablation studies.

\subsection{From BasicTAD to PlusTAD}
\label{frombasictadtobasictadplus}
Based on our modular TAD framework, we perform comprehensive studies of the basic options and come up with a simple yet effective TAD baseline, termed BasicTAD.
Furthermore, based on the BasicTAD, we introduce three customized improvements to fully unleash the power of this simple and end-to-end detection pipeline, and the upgraded framework is called \textbf{PlusTAD}.
Specifically, we propose two network structure designs, Temporal Preservation and Spatial Preservation, to improve the quality of the temporal features of networks' backbone and neck parts.
We further study and adopt stronger data augmentations for training and multi-view ensemble methods in the testing phase.

\subsubsection{Temporal Preservation for Backbone}
\label{temporalpreservationforbackbone}
The existing TAD methods use backbones that downsample temporal and spatial information and require dense frames sampled at high FPS as input.

Intuitively, denser inputs generally lead to better detection performance. To alleviate the huge computational overhead imposed by dense inputs, the existing practice is to perform temporal downsampling on the inputs in the early stages of the backbone. However, our subsequent experiments show that this approach does not always yield gains. 
Due to insufficient discriminative power between adjacent frames in shallow layers, it is hard for the early stages of the backbone to learn how to extract useful temporal signals from these subtle frame changes. Therefore, we propose \textbf{Temporal Preservation} (termed as ``TP'' for short) design that feeds frames into the model from the beginning at an explicitly sampled frame rate (sparse or dense) without downsampling in the backbone.
For long actions in datasets like THUMOS14, sparse input without temporal downsampling in the backbone is enough to predict actions. Though sparse input may not be suitable for fine-grained action detection, maintaining the temporal dimension and adopting dense input can encourage the backbone to capture their subtle changes. TP design is effective and helpful for both scenarios above.

\subsubsection{Spatial Preservation for Neck}
\label{spatialpreservationforneck}
Many existing TAD works~\citep{afsd,rgb_enough} squeeze the spatial dimension of features before feeding them into the neck module. 
We argue that the squeezing operation prematurely drops the spatial dimension, making the feature pyramid constructed by the neck module unable to capture spatial-sensitive multi-scale features.

In our further experiments, we find it is worth preserving spatial dimension. 
So we propose a design named \textbf{Spatial Preservation} (termed as ``SP'' for short) that postpones the spatial squeezing until after the neck and replaces the 1D operators with 3D operators. 
It introduces the local spatial context and enhances the robustness of multi-scale spatiotemporal information. 
We discuss its performance improvement in the experimental section.

\subsubsection{Enhanced Data Augmentation} \label{detaileddataaugmentation}
In Section~\ref{sec:dataaugmentation}, we equip BasicTAD with various image data augmentations in the training phase.
To further improve the performance of our PlusTAD pipeline, we explore more data augmentation methods. Through ablation experiments, we investigate the effects of more different spatial resolutions on TAD performance in the training stage.

Furthermore, considering the data augmentation in the testing phase mainly works on the temporal and spatial levels, we adopt various test augmentation methods to enhance the robustness of the predictions. 
In detail, we employ two spatial-level data augmentation methods for each temporal window, namely ``ThreeCrop'' and ``Flip'', individually or simultaneously. 
We also use the reverse sliding window, termed ``Backward'', to increase the density of temporal windows and combine the predictions of each temporal window for post-processing.

\subsection{Training and Testing}
\subsubsection{Training} 
In the training phase, we randomly sample a fix-sized temporal window of consecutive frames in each untrimmed video per iteration and feed them into our model.
We train BasicTAD and PlusTAD with both heads as a multi-task loss function $L$, including a classification loss  $L_{\rm cls}$, a regression loss $L_{\rm reg}$.
\begin{equation}
L=L_{\rm cls}+\alpha L_{\rm reg},
\end{equation} 
where $\alpha$ is a hyper-parameter to balance these two terms and set them to 1. classification loss: We use focal loss~\citep{focalloss} as classification loss for its ability to solve the problem of positive and negative sample imbalance. The formal expression of focal loss is as follows:
\begin{equation}
L_{\rm cls}=-(1-p_t)^{\gamma}log(p_t)
\end{equation} 
where $p_t$ represents the probability of each action category after softmax, and $\gamma$ represents the modulation factor that focuses on hard samples. 
Meanwhile, Distance-IoU (DIoU) loss~\citep{diou} is adopted as regression loss for faster convergence during training and more accurate boundary regression. The formal expression of DIoU loss is as follows:
\begin{equation}
L_{\rm reg}=1-\left |  \frac{B\cap B^{gt}}{B \cup  B^{gt}}  \right | + \frac{\rho^{2}(b,b^{gt})}{c^{2}} 
\end{equation} 
where $B$ represents the boundary of the proposed action, $B^{gt}$ represents the boundary of ground truth action, $\rho^{2}(b,b^{gt})$ represents the square of the Euclidean distance between the center points of $B$ and $B^{gt}$ and $c$ represents the shortest length that simultaneously covers $B$ and $B^{gt}$.

\subsubsection{Testing}
We use sliding windows with overlap to sample fixed-number frames and feed them into BasicTAD and PlusTAD in the testing stage. 
Our model outputs $\{(s_i,e_i,p_i)\}_{i=1}^{m}$ as the predicted action set, where $i$ and $m$ is the $i$-th action and total number of predicted action. $s_i$, $e_i$ represent the start and end time of the $i$-th action $i$ and $p_i$ is its score. 
We adopt NMW~\citep{nmw} for both heads to remove the redundant action segments.

\section{Experiments}

\begin{table}[t]
\centering
\small
\caption{\textbf{Summary of three datasets.} Video number, category number and other information about THUMOS14, FineAction and ActivityNet-v1.3 are listed in this table.}
\resizebox{0.5\textwidth}{!}{
\begin{tabular}{cccccc}

\toprule 
Dataset & Video & Category & Instance & Duration & Type  \\ \midrule
THUMOS14    &  413 & 20 &6,316 & 4.3 s & sports    \\
FineAction    &  16,732 & 106 &103,324 & 7.1 s & daily events  \\
ActivityNet-v1.3     &  19,994 & 200 &23,064 & 49.2 s & daily events          \\\bottomrule
\end{tabular}
}
\label{table:dataset_info}
\end{table}

\begin{figure}[!t]
  \includegraphics[width=0.5\textwidth]{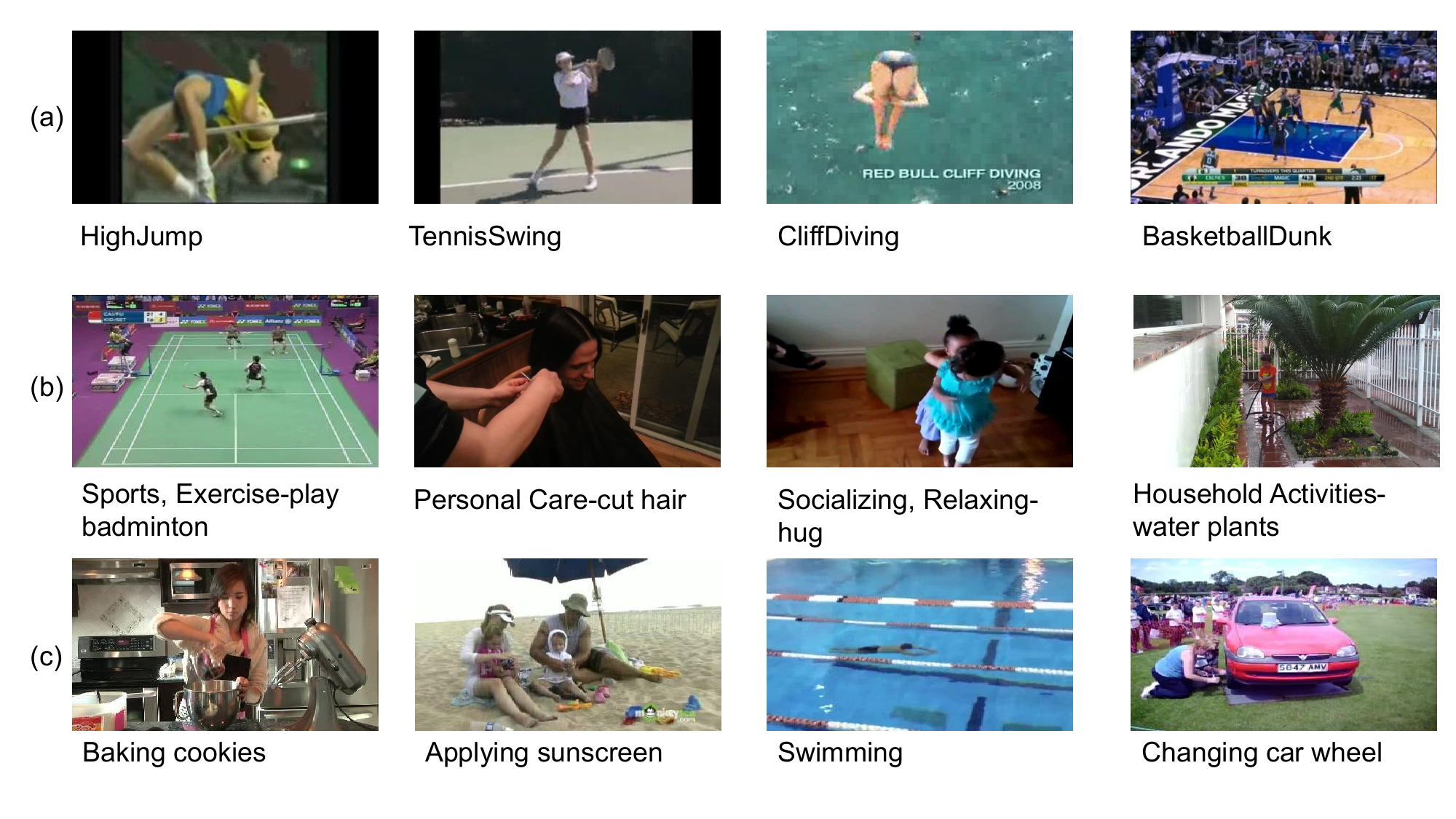}
  \caption{
\textbf{Samples of three TAD datasets.} (a): A few examples in THUMOS14. All actions are sports-related. (b): A few examples in FineAction. These samples' categories are shown in the form of ``top-level category"-``bottom-level category". Its action scenes are more varied than THUMOS14. (c): A few examples in ActivityNet-v1.3. It has plenty of action scenes. Unlike FineAction, actions in ActivityNet-v1.3 are mostly long-term actions.
}
  \label{fig:cviu_dataset}
\end{figure}

\subsection{Datasets}
We perform extensive experiments on three datasets, summarized in Fig~\ref{fig:cviu_dataset} and Table~\ref{table:dataset_info}, to demonstrate the effectiveness of our BasicTAD and PlusTAD. 
\textbf{THUMOS14}~\citep{THUMOS14} is a commonly-used dataset in TAD, containing 200 validation videos and 212 test videos with labeled temporal annotations from 20 action categories in sports. 
\textbf{FineAction}~\citep{fineaction} is a newly collected large-scale fine-grained TAD dataset containing 57,752 training instances from 8,440 videos and 24,236 validation instances from 4,174 videos and 21,336 testing instances from 4,118 videos. It contains 106 action categories within a new taxonomy of three-level granularity. This taxonomy consists of 4 top-level categories, 17 middle-level categories, and 106 bottom-level categories. The 4 top-level categories are ``Household Activities'', ``Personal Care'', ``Socializing, Relaxing'' and ``Sports, Exercise''. With richer action scenes and finer-grained action categories, FineAction will be a challenging data set in the TAD task.
\textbf{ActivityNet-v1.3}~\citep{anet} is a large-scale dataset containing 10,024 training videos, 4,926 validation videos, and 5,044 test videos belonging to 200 activities covering sports, household, and working actions. Different from THUMOS14 and FineAction, which consist mostly of short actions, most videos in ActivityNet-v1.3 contain only one long-term action instance, and the action frames exceed 64\% of all frames. 

\subsection{Evaluation Metric}
Following previous work, we report the \textbf{mean average precision (mAP)} with tIoU thresholds $[0.3:0.1:0.7]$ on the test set of THUMOS14.
And on the validation set of FineAction and ActivityNet-v1.3, the thresholds are set as $[0.5,0.05,0.95]$.
We use ``$\text{Avg}$'' to represent the average mAP on THUMOS14, FineAction, and ActivityNet-v1.3.

\noindent 
\subsection{Implementation Details}
We perform experiments using BasicTAD and PlusTAD on THUMOS14~\citep{THUMOS14}. Though THUMOS14 is a scene-biased dataset that can not reflect the importance of temporal dynamics in classification, it can provide temporal dynamics for the regression of action boundaries. This means that the conclusions obtained from the ablation experiments on THUMOS14 are universal to a certain extent, so we choose THUMOS14 to do ablation studies. 
We sample temporal windows of 32 seconds for both configurations, covering over 99.7$\%$ action instances. 
We use 768 frames at 24FPS for BasicTAD and 96 frames at 3FPS for PlusTAD. 
In the training phase, we resize the \textbf{original size} (the size of the original frames) to $128\times128$ for BasicTAD and short-128 (the short side of the frame is set to 128) for PlusTAD. 
We randomly sample a temporal window in each untrimmed video per iteration.
We set the \textbf{crop size} (the size of the cropped images) to $112\times112$. 
The photo distortion and random spatial rotation are consistent with~\citep{rgb_enough,ssd}.
In the testing phase, we sample uniform sliding windows whose sampling stride between adjacent sliding windows is set to 25\% of the window length. 
And the center crop of $112\times112$ is used. 
If there is no special emphasis, the above configuration will be the default configuration for subsequent experiments based on THUMOS14.

We use PlusTAD to perform further experiments on FineAction~\citep{fineaction} and ActivityNet-v1.3~\citep{anet}.
On FineAction, we sample RGB frames at 2FPS. 
In the training phase, we set the original size to short-256 and the crop size to $224\times224$.
In the testing phase, we use the center crop of $224\times224$.
On ActivityNet-v1.3, we follow the settings in AFSD~\citep{afsd} that we sample frames at different fps and ensure the number of frames in each video is the same. We set the original size to $224\times224$.

We use SlowOnly~\citep{slowfast} pre-trained on Kinetics~\citep{kinetics} as our backbone by default, where all batch normalization layers are frozen.
We train the model using SGD with a momentum of 0.9 and weight decay of 0.0001. The batch size is set as 16.
The learning rate schedule is annealing down from 0.01 to 0.0001 every 1200 iterations on THUMOS14, 20 epochs on FineAction, and ActivityNet-v1.3 using the cosine decay rule.

\subsection{Ablation Studies on BasicTAD} 

We first begin the ablation study on the design of each component in our proposed modular TAD framework. In the previous section, we introduced the basic and simple options for each step in our modular framework. In this section, we will perform extensive studies over these basic designs through in-depth and step-by-step ablation experiments. Our goal is to discover a simple yet must-known TAD baseline.

\begin{table*}[t]
\centering
\small
\caption{\textbf{Performance of BasicTAD on different backbones on THUMOS14~\citep{THUMOS14}}. We replace SlowOnly with C3D~\citep{c3d}, I3D~\citep{i3d} and R50-I3D~\citep{r50-i3d-non-local} for our BasicTAD to perform experiments, where I3D only retains RGB branch. }
\label{table:backbone}
\setlength\tabcolsep{4.7mm}
\resizebox{1.0\textwidth}{!}{
\begin{tabular}{l|c|cccccc}
\toprule

Method& Backbone & mAP@0.3 & mAP@0.4 & mAP@0.5 & mAP@0.6 & mAP@0.7 & Avg \\
\midrule
\multirow{4}{*}{Anchor-based}&C3D&54.4 &50.8 &45.7 & 37.4& 26.1&42.9\\
~&I3D& 59.5& 56.0&51.4 &41.8 &28.3 &47.4\\
~&R50-I3D& 62.8& 59.5&53.8 &43.6 &30.1 &50.0\\
~&SlowOnly ($\times8$)&\textbf{63.1} &\textbf{59.5} & \textbf{54.3}&\textbf{43.6} &\textbf{30.5} &\bf{50.2}\\
\midrule
\multirow{4}{*}{Anchor-free}&C3D&56.8 & 50.1& 44.7& 35.2&24.3&42.2\\
~&I3D& 61.7& 56.9& 49.3&38.7 &26.1 &46.6\\
~&R50-I3D& \textbf{63.7}& 58.5& 51.6&41.0 &30.8 &49.1\\
~&SlowOnly ($\times8$)&63.2 &\textbf{59.7} &\textbf{52.6} &\textbf{44.5} &\textbf{32.7} &\bf{50.5}\\
\bottomrule
\end{tabular}
}
\end{table*}

\begin{table*}[t]
\centering
\small
\caption{\textbf{Study on the effectiveness of end-to-end training based on the anchor-based BasicTAD with different backbones on THUMOS14~\citep{THUMOS14}}. ``e2e'' is short for end-to-end training. We freeze all layers in the backbone to construct a non-end-to-end training manner.}
\label{table:e2e}
\setlength\tabcolsep{5mm}
\resizebox{1.0\textwidth}{!}{
\begin{tabular}{cc|cccccc}
\toprule
     Backbone                     & e2e & mAP@0.3 & mAP@0.4 & mAP@0.5 & mAP@0.6 & mAP@0.7 & Avg  \\ 
     \midrule
\multirow{2}{*}{C3D}    & \CheckmarkBold    & \textbf{54.4}    & \textbf{50.8}    & \textbf{45.7}    & \textbf{37.4}    & \textbf{26.1}    & \textbf{42.9} \\
           &  \XSolidBrush   & 32.0    & 27.2    & 22.0    & 14.4    & 7.6     & 20.6 \\ 
           \midrule
          \multirow{2}{*}{I3D}     &  \CheckmarkBold   & \textbf{59.5}    & \textbf{56.0}    & \textbf{51.4}    & \textbf{41.8}    & \textbf{28.3}    & \textbf{47.4} \\
                     &  \XSolidBrush   & 35.2    & 29.7    & 23.1    & 15.9    & 8.1     & 22.4 \\ 
                     \midrule
              \multirow{2}{*}{R50-I3D} &  \CheckmarkBold   & \textbf{62.8}    & \textbf{59.5}    & \textbf{53.8}    & \textbf{43.6}    & \textbf{30.1}    & \textbf{50.0} \\
                 & \XSolidBrush    & 36.5    & 31.8    & 26.7    & 19.6    & 12.3    & 25.4 \\ \bottomrule
              \multirow{2}{*}{SlowOnly ($\times8$)} &  \CheckmarkBold   & \textbf{63.1}    & \textbf{59.5}    & \textbf{54.3}    & \textbf{43.6}    & \textbf{30.5}    & \textbf{50.2} \\
                 & \XSolidBrush    &  37.3   &  32.4   &  26.6   & 19.8   &  13.0   & 25.8 \\ \bottomrule
\end{tabular}
}
\end{table*}

\noindent
\subsubsection{Study on different backbones}
The first and most important choice in our modular TAD framework is the backbone design.
The previous works~\citep{rgb_enough,r-c3d,pbrnet,afsd,sparse-rcnn-tad} with end-to-end training manner usually input many frames and use backbones with $8\times$ temporal downsampling, such as C3D~\citep{c3d}, I3D~\citep{i3d}, and R50-I3D~\citep{r50-i3d-non-local}.
This method can increase the utilization of the data source as much as possible. Meanwhile, using $8\times$ spatiotemporal downsampling can reduce the computational overhead. 
However, these backbones are relatively outdated and inferior to the recent SlowOnly backbone~\citep{slowfast}. To fully unleash the power of our proposed modular TAD framework with end-to-end training, we also try SlowOnly backbone in our TAD framework.
For a fair comparison with previous backbones, we align with the settings of these previous works, and $8\times$ temporal downsampling is adopted. 
Hence, we insert a $2\times$ downsampling layer before res-2, res-3, and res-5(res-x denotes the x-th res-stage in SlowOnly) of the SlowOnly backbone, respectively. 

Inputting \textbf{768 frames at 24FPS} under end-to-end training, we compare the performance of BasicTAD, which uses C3D, I3D, R50-I3D, and SlowOnly with $8\times$ downsampling in Table~\ref{table:backbone}.
These four backbone encoders are all 3D CNN methods that capture spatiotemporal information between frames by performing 3D convolution. 
As shown in Table~\ref{table:backbone}, BasicTAD with C3D does not perform well due to its limited representation power caused by relatively shallow networks. 
Meanwhile, I3D performs worse than SlowOnly with $8\times$ downsampling. This can imply that the ResNet50~\citep{resnet} network outperforms the Inception-v1~\citep{bn-inception} network in the TAD task.
R50-I3D is slightly weaker than SlowOnly with $8\times$ downsampling because they both are based on ResNet. R50-I3D performs $8\times$ downsampling at the beginning of the backbone, while SlowOnly with $8\times$ downsampling delays the timing of backbone downsampling. This difference will contribute to the slight performance difference. This downsampling location will be analyzed in the next section.

\subsubsection{Study on the gain of end-to-end training}
In this section, we study the gain of end-to-end training BasicTAD with different backbones. 
Table~\ref{table:e2e} shows that abandoning the end-to-end training strategy leads to significantly worse detection results for the anchor-based method across all four backbones. 
While this method reduces training costs, it fails to fully leverage the modeling capabilities of the backbone, given the gap between TAD and action recognition tasks.
This underscores the importance of a trainable backbone in the TAD pipeline.

\begin{table*}[t]
\centering
\small
\caption{\textbf{Comparison between different locations of downsampling in BasicTAD on THUMOS14~\citep{THUMOS14}}.
We choose three of four intervals between these res-layers to insert the max-pool downsampling operation.
In this table, choosing an interval before a res-layer will be indicated by a tick.
}
\label{table:basicTAD-downsample}
\setlength\tabcolsep{2.4mm}
\resizebox{1.0\textwidth}{!}{
\begin{tabular}{l|cccc|c|cccccc}
\toprule
Method      & res-2 & res-3 & res-4 & res-5 & FLOPs& mAP@0.3 & mAP@0.4 & mAP@0.5 & mAP@0.6 & mAP@0.7 & Avg \\ \midrule
 \multirow{4}{*}{Anchor-based}  & \checkmark& \checkmark &  \checkmark &    &  227.7G  &  55.9   &   52.5  &   47.3  &  40.3   &  29.2  &  45.1\\
   &  \checkmark    &    \checkmark    &        &    \checkmark    &  280.2G  & \textbf{63.1}    & \textbf{59.5}    & \textbf{54.3}    & \textbf{43.6}    & \textbf{30.5}    &{\bf 50.2}  \\
     &  \checkmark    &        &     \checkmark   &    \checkmark    &   329.9G  &55.3     &51.5     &  46.8   &38.0    &  26.9  & 43.7\\
&   &  \checkmark   &     \checkmark   &    \checkmark    &  395.4G   &56.5     &53.1     &48.2     &40.1     &29.0    & 45.4\\ \midrule
      \multirow{4}{*}{Anchor-free} &    \checkmark    &     \checkmark   &   \checkmark     &        & 245.8G  &  59.2   &   55.0  &   47.5  &  37.2   &  27.3  &  45.3 \\
    &     \checkmark   &    \checkmark    &        &    \checkmark    & 298.4G &    63.2  &   \textbf{59.7}  &   \textbf{52.6}  &  \textbf{44.5}   &  \textbf{32.7}  &  {\bf 50.5} \\
     &     \checkmark   &        &    \checkmark    &     \checkmark   & 348.1G &   63.8   &   58.2  &   50.4  &  41.8   &  30.1  &  48.9\\
      &        &     \checkmark   &    \checkmark    &    \checkmark    & 413.5G &    \textbf{64.6}   &   58.8  &  51.4   &  42.6   &  31.2  &  49.7\\
\bottomrule
\end{tabular}
}
\end{table*}

\begin{table*}[t]
\centering
\small
\caption{\textbf{Comparison between three kinds of the neck on THUMOS14~\citep{THUMOS14}}. We test three different kinds of neck and compare the results on THUMOS14. We denote the down-sampling operator of Lateral-TFPN as ``-'' because it uses the multi-scale feature from the backbone.}
\label{table:neck_operator}
\setlength\tabcolsep{3.3mm}
\resizebox{1.0\textwidth}{!}{
\begin{tabular}{l|c|c|cccccc}
\toprule
Method  & Neck                    & Operator & mAP@0.3 & mAP@0.4 & mAP@0.5 & mAP@0.6 & mAP@0.7 & Avg  \\ \midrule
\multirow{5}{*}{Anchor-based}   & Lateral-TFPN      & -     & 59.1    & 55.4    & 50.0    & 40.9    & 27.9    & 46.7 \\
 ~   & Post-TDM      & Conv     & 56.7    & 53.1    & 48.0    & 38.8    & 27.0    & 44.7 \\
  ~   & Post-TDM      & Maxpool  & 54.1    & 52.0    & 46.3    & 37.9    & 27.8    & 43.4 \\
 ~   & Post-TDM-TFPN & Conv     & \textbf{63.1}    & \textbf{59.5}    & \textbf{54.3}    & \textbf{43.6}    & 30.5    & \bf{50.2} \\
 ~   & Post-TDM-TFPN & Maxpool  & 58.7    & 55.2    & 50.7    & 42.0    & \textbf{30.6}    & 47.5 \\ \midrule
\multirow{5}{*}{Anchor-free}  & Lateral-TFPN      & -     & 56.1    & 50.2    & 43.6    & 34.1    & 24.5    & 41.7 \\
 ~   & Post-TDM      & Conv     & \textbf{63.9}    & 58.7    & 50.2    & 40.0    & 29.1    & 48.4 \\
  ~   & Post-TDM      & Maxpool  & 63.2    & \textbf{59.7}    & \textbf{52.6}    & \textbf{44.5}    & \textbf{32.7}    & \bf{50.5} \\
 ~   & Post-TDM-TFPN & Conv     & 62.3    & 56.5    & 49.1    & 38.3    & 24.6    & 46.2 \\
~& Post-TDM-TFPN & Maxpool  & 61.8    & 56.6    & 49.2    & 39.9    & 28.7    & 47.2 \\ \bottomrule
\end{tabular}
}
\end{table*}

\noindent
\subsubsection{Study on Downsampling Locations in backbone}
Following the previous section, we further study downsampling locations in the SlowOnly backbone.
To construct a structure with $8\times$ temporal downsampling, the backbone of BasicTAD need to choose three different places to operate $2\times$ downsampling.
Since R50-based SlowOnly has five stages, we can choose three of four intervals between these res-stages. In this sense, we can downsample the spatiotemporal features before res-2, res-3, res-4, and res-5 (res-$x$ denotes the $x$-th res-stage in SlowOnly). Inserting downsampling layers at different locations can lead to differences in model performance due to different temporal receptive fields and aggregations. Meanwhile, their corresponding computing costs will be different as well.

Following the input in the previous section, we compare the performance of BasicTAD with different downsampling locations. The results are shown in Table~\ref{table:basicTAD-downsample}. 
In the backbone's first half stages (res-2 and res-3), the features are highly redundant in the temporal dimension. If we put the temporal downsampling layers at the first three stages, it can save the computational cost most but also achieves the worst detection mAP. However, if we simply change a single downsampling location from res-4 to res-5, it leads to a large performance improvement of around 5\% mAP. Meanwhile, we notice that other configurations of downsampling locations achieve a weaker performance than the previous version yet with higher computational costs. 
From the above results, it can be found that the temporal features are highly redundant in the first half stages, and the latter half stages are key stages of feature encoding, so temporal downsampling cannot be easily performed there.
Thus, we keep the temporal downsampling locations before res-2, res-3, and res-5 by default.

\subsubsection{Study on Neck Design} 
After the ablation studies on the backbone design, we now turn to the exploration of the neck design.
Based on the best result in Table~\ref{table:basicTAD-downsample}, we set the down-sampling locations at res-2, res-3, and res-5 in this ablation study.
Specifically, we compare three kinds of neck modules for anchor-based and anchor-free TAD methods, namely Lateral-TFPN, Post-TDM, and Post-TDM-TFPN, as introduced above. These neck modules with different down-sampling operators yield multi-resolution representations for action detection. The results are listed in Table~\ref{table:neck_operator}.
As shown in the table, Lateral-TFPN achieves the worst performance because the feature maps lack enough high-level semantic information in the early stages of the backbone.
In order to keep high-level semantic information in the neck module, it is necessary to build the neck representation right behind the backbone.

For Post-TDM and Post-TDM-TFPN, we explore two basic operators in the downsampling procedure, namely convolution, and max-pool. As demonstrated in Table~\ref{table:neck_operator}, Post-TDM with the max-pool operator and Post-TDM-TFPN with the convolution operator is more suitable for the anchor-free method and the anchor-based method, respectively.
This difference is due to the different detection and training mechanisms of the anchor-free method and the anchor-based method.
The anchor-based TAD method adopts the global sample assignment mechanism based on tIoU. 
It depends on the global multi-scale context aggregated by Post-TDM-TFPN with convolution operations. 
Instead, the anchor-free TAD method uses a sample assignment mechanism based on the multi-scale center points, so it is subject to local high-frequent boundary information in the temporal dimension to directly regress the locations of action boundaries.

Based on the above results and analysis, we apply Post-TDM with the max-pool operator and Post-TDM-TFPN with the convolution operator to the anchor-free method and the anchor-based method in BasicTAD.

\noindent
\subsubsection{Study on Head Design} 
To ensure the simplicity principle in BasicTAD, we introduce basic one-stage anchor-based and anchor-free methods in our study.
In the previous ablation experiments on backbone and neck design for BasicTAD, we conducted experiments with both detection heads. The results are shown in Table~\ref{table:backbone}, Table~\ref{table:basicTAD-downsample}, and Table~\ref{table:neck_operator}. From these results, we can also provide some comparative analysis for head design.

As shown in Table~\ref{table:basicTAD-downsample}, the anchor-based TAD method has slightly lower FLOPs than the anchor-free TAD method due to its smaller number of channels in the detection head. Although the max-pool operator does not contain any learnable parameters, the anchor-free method does not reduce the dimension of the last layer features from SlowOnly, which is 2048. It leads to more FLOPs compared with the anchor-based method. For detection accuracy, the anchor-free TAD method achieves a slightly better mAP than the anchor-based TAD method.

In our design, the anchor-based and anchor-free BasicTAD are both dense detectors.
As dense prediction methods, anchor-based and anchor-free heads rely on post-processing to suppress redundant predictions to yield the final detection results.
In our BasicTAD, we generally follow the box suppression methods in object detection. Specifically, various classical  suppression algorithms, \emph{i.e.}, Non-Maximum Suppression (NMS)~\citep{nms}, Non-Maximum Weighting (NMW)~\citep{nmw} are introduced to suppress bounding boxes in post-processing. 
We perform a comparative study on them to determine the best option for anchor-free and anchor-based TAD methods.

The experiment results on the choice of the post-processing algorithms are reported in Table~\ref{table:nms}.
For the anchor-free method, the detection result is not sensitive to the choice of the post-processing algorithm, and using NMW is slightly better. 
However, the results of the anchor-based method are quite different, where the choice of the post-processing algorithm significantly affects the detection results.
One possible reason is that in TAD tasks, annotations tend to be very sparse, and the anchor-based method generates too many redundant predictions that we need an advanced suppression algorithm. 
NMW performs better on anchor-based and anchor-free methods, so we use NMW in the subsequent experiments.

\begin{table*}[t]
\centering
\small
\caption{\textbf{Comparison of Post-Processing strateg on THUMOS14~\citep{THUMOS14}}. We take both Non-Maximum Suppression (NMS) and Non-Maximum Weighting (NMW) into account for both anchor-free and anchor-based methods. NMW performs better than NMS for its weighted average mechanism for redundant accurate boundaries. }
\label{table:nms}
\setlength\tabcolsep{4.6mm}
\resizebox{1.0\textwidth}{!}{
\begin{tabular}{l|c|cccccc}
\toprule
Method  & Post Processing & mAP@0.3 & mAP@0.4 & mAP@0.5 & mAP@0.6 & mAP@0.7 & Avg  \\ \midrule
\multirow{2}{*}{Anchor-based}           & NMS                     &   61.4      &  57.8      &    51.8     & 42.1      &  \textbf{30.6}     &   48.7   \\
   ~       & NMW                     & \textbf{63.1}    &  \textbf{59.5}  &  \textbf{54.3}   &  \textbf{43.6}   &  30.5   &  \textbf{50.2}\\ \midrule
\multirow{2}{*}{Anchor-free}          & NMS  &  63.2   &  59.7   &  52.6   &   \textbf{44.5}  &  32.7   & 50.5 \\
   ~         & NMW  &\textbf{64.7}     &  \textbf{60.0}   &  \textbf{52.8}   &  43.7   & \textbf{33.0}   &  \textbf{50.8}\\ \bottomrule
\end{tabular}
}
\end{table*}

\begin{table*}[t]
\centering
\small
\caption{\textbf{Ablation study on improvements of PlusTAD on THUMOS14~\citep{THUMOS14}}. Temporal Preservation is for backbone design (indicated as ``TP''), and Spatial Preservation is for neck design (indicated as ``SP''). }
\label{table:xx}
\setlength\tabcolsep{2.55mm}{
\resizebox{1.0\textwidth}{!}{
\begin{tabular}{l|cc|cc|cccccc|c}
\toprule
Method & TP&SP& FPS& Frames & mAP@0.3 & mAP@0.4 & mAP@0.5 & mAP@0.6 & mAP@0.7 & Avg & FLOPs  \\ \midrule
\multirow{5}{*}{Anchor-based}&  &  &24& 768 &63.1 &59.5 & 54.3 &43.6 &30.5&50.2&280.2G        \\ 
& \checkmark &   &3& 96 &63.9 &59.4 & 53.7 &44.3 &31.0  &50.5 &133.3G       \\

& \checkmark & \checkmark &3 &96 &63.2 &59.7 & 54.4 &45.6 &32.2  &51.0 &136.4G       \\

& \checkmark &   &6& 192 &\textbf{65.4} &61.8 & 56.4 &47.9 &33.8  &53.1 &266.0G       \\ 

& \checkmark & \checkmark &6& 192 &65.1 &\textbf{62.0} & \textbf{57.1} &\textbf{48.9} &\textbf{33.9}  & {\bf 53.4} &272.9G       \\ 

\midrule
\multirow{5}{*}{Anchor-free}&  &  &24& 768 &64.7 & 60.0 & 52.8 &43.7 &33.0&50.8&298.4G        \\ 
& \checkmark &   &3& 96 &65.7 &60.7 & 52.8 &41.7 &29.3  &50.0 &151.4G       \\ 
& \checkmark & \checkmark &3& 96 & 67.3 & 61.6 & 54.4 & 41.2 & 29.5 & 50.8  &151.5G       \\
& \checkmark &   &6& 192 &67.5 &62.6 & 55.4 &\textbf{45.7} &\textbf{33.7}  &53.0 &  284.2G     \\ 
& \checkmark & \checkmark &6& 192 &\textbf{69.9} &\textbf{64.3} &\textbf{56.8} &44.7 & 32.1 & {\bf 53.6} &284.2G       \\ 
\bottomrule
\end{tabular}
}
}
\end{table*}

\subsection{Ablation Studies on PlusTAD}
We have explored several basic settings of each module in our modular pipeline and come up with a very simple TAD baseline method, termed BasicTAD.
These experiments are all based on the basic design described in Section~\ref{basictad}.
In this subsection, we further perform ablation experiments to validate the improvements proposed over the BasicTAD design. Specifically, we provide empirical results for the three improvements, namely temporal preservation of backbone design, spatial preservation of neck, and more training augmentations. With these three improvements, we obtain another more powerful TAD baseline method, termed as PlusTAD.

\subsubsection{Temporal Preservation in Backbone Design}

Temporal sampling is an important but less studied factor influencing the design of the TAD method. We have presented a Temporal Preservation (TP) principle in our PlusTAD design as we need to keep as much temporal semantic information as possible for efficient localization of each action instance. 
We conduct the following experiments to study the effectiveness of TP for the backbone design.
In detail, we directly remove $8\times$ temporal downsampling in the backbone design. Thus the temporal size of feature maps keeps the same with the input frames for all stages in ResNet. In order to keep the final temporal feature the same as the BasicTAD, we use {\bf 96 frames at 3FPS} to replace {\bf 768 frames at 24FPS} in BasicTAD. We change the sampling FPS to keep the temporal window size the same. 

As is shown in Table~\ref{table:xx}, compared with BasicTAD (the first row in the table), using a video clip of 96 frames at 3FPS achieves a similar performance for both anchor-based and anchor-free TAD methods. However, the computation overhead is only half of BasicTAD, indicating that our proposed TP principle in backbone design is a good practice for increasing TAD running speed without mAP loss. 
To fully unleash the modeling power of our TP-equipped backbone in TAD, we further increase the sampling FPS and sampling frame number to provide richer information.
When using video clips of 192 frames at 6FPS, the average mAP of the anchor-based and the anchor-free methods is improved by 2.9 and 2.2 over the BasicTAD, respectively. But, their overall FLOPs are still lower than that of BasicTAD.
This performance demonstrates that preserving the temporal resolution in the backbone design makes it more effective and efficient in capturing temporal information. Therefore, in our PlusTAD, we will employ the TP principle in our design by default.

\subsubsection{Spatial Preservation in Neck Design}

Our proposed Spatial Preservation (SP) design aims to build spatial-sensitive temporal multi-scale features in the neck module. 
We maintain the spatial dimension of features in the neck module and perform the spatial squeezing after completing the multi-scale feature construction.

As shown in Table~\ref{table:xx}, with the same other settings, using SP for the neck module with 96 and 192 frames can obtain 0.8 and 0.6 mAP improvement in the anchor-free method, and 0.5 and 0.3 mAP improvement in the anchor-based method, respectively. 
This is due to the fact that additional spatial dimension can make features capture rich structure on the spatial locations of actions occurring at different scales. Meanwhile, we observe that the computational cost of PlusTAD with SP is almost the same as without SP. The performance of SP demonstrates that it is a simple module for improving temporal action detection performance. Therefore, by default, we incorporate the SP design principle in our PlusTAD.

\begin{table*}[t]
\centering
\small
\caption{\textbf{Comparison of the different spatial sizes in the training stage on THUMOS14~\citep{THUMOS14}}.
Original size means input frame size, and the crop size indicates the cropped patch resolution as the network input. 
}
\label{table:BasicTAD Plus-origin-spatial-size}
\setlength\tabcolsep{2.65mm}
\resizebox{1.0\textwidth}{!}{
\begin{tabular}{l|c |c|ccccc c|c}
\toprule
Method&Original size& Crop size & mAP@0.3 & mAP@0.4 & mAP@0.5 & mAP@0.6 & mAP@0.7&$\text{Avg}$  & FLOPs \\
 \midrule
\multirow{4}{*}{Anchor-based}&$128\times128$  & $112\times112$&63.2&59.7&54.4&45.6&32.2&51.0 & 136.4G\\
~&$171\times128$  &$112\times112$ &66.8&63.3&57.8&48.8&32.9&53.9  & 136.4G \\
~&short-128  & $112\times112$&68.4&65.0&58.6&49.2&33.5&54.9  & 136.4G \\
~&short-180 &$160\times160$&\textbf{71.2}&\textbf{66.8}&\textbf{61.2}&\textbf{50.1}&\textbf{36.3}& \textbf{57.1} & 262.1G    \\
\midrule
\multirow{4}{*}{Anchor-free}&$128\times128$  & $112\times112$&67.3 &61.6 &54.4 &41.3 &29.5 &50.8 &151.5G \\
~&$171\times128$  &$112\times112$& 68.1&64.3&55.2 &45.3 &31.9 &53.0  &151.5G  \\
~&short-128  & $112\times112$ &70.4 & 65.5&57.6 &46.0 &33.2 &54.5 &151.5G\\
~&short-180 &$160\times160$ &\textbf{72.5} &\textbf{66.8} & \textbf{59.1}& \textbf{48.4} &\textbf{35.0} &  \textbf{56.4}  &275.9G\\
\bottomrule
\end{tabular}
}
\end{table*}

\begin{table*}[t]
\centering
\small
\caption{\textbf{Ablation study results on testing augmentation on THUMOS14~\citep{THUMOS14}.} 
For ThreeCrop and Flip augmentation, we fuse the features after the neck. For backward augmentation, we fuse the detection results in the post-processing phase. 
}
\label{table:test-augmentation-method}
\setlength\tabcolsep{2.8mm}
\resizebox{1.0\textwidth}{!}{
\begin{tabular}{l|l|c |cccccc}
\toprule
 Method& Test Aug & Crop Size & mAP@0.3 & mAP@0.4 & mAP@0.5 & mAP@0.6 & mAP@0.7& $\text{Avg}$ \\
 \midrule
\multirow{5}{*}{Anchor-based}&CenterCrop    & $160\times160$  &  71.2&	66.8&	61.2&	50.1&	36.3&	57.1   \\

~&CenterCrop    & $180\times180$  &  71.6&	67.3&	61.6&	50.8&	34.7	&  57.2  \\

~&CenterCrop+Backward    & $180\times180$  &  71.1&	67.4&	61.1&	51.0&	34.5 &  57.0
\\

~&ThreeCrop    & $180\times180$  &  71.7&	\textbf{67.9}&	62.0	&50.7&	\textbf{35.6}&	 {\bf 57.6}  \\

~&CenterCrop+Flip    & $180\times180$  &  \textbf{71.9}&	67.7&	\textbf{62.1}&	\textbf{51.0}&	35.2& {\bf 57.6}   \\
\midrule
\multirow{5}{*}{Anchor-free}&CenterCrop    & $160\times160$  & 72.5 & 66.8& 59.1&48.4 &35.0		& 56.4  \\

~&CenterCrop    & $180\times180$  & 72.9&66.3	& 59.5&48.2	 &	 35.1	&  56.4  \\

~&CenterCrop+Backward    & $180\times180$  & \textbf{72.9} &	66.7&	\textbf{59.9} &	47.6 &	33.9&  56.2 \\

~&ThreeCrop    & $180\times180$  & 72.3  &	\textbf{67.6} &	  59.0	& 48.5&\textbf{35.9} &	 {\bf 56.7} \\

~&CenterCrop+Flip    & $180\times180$  & 72.4 &66.8	&59.7	&\textbf{48.9} &35.0&  56.6\\
\bottomrule
\end{tabular}
}
\end{table*}

\subsubsection{Effectiveness of Training Augmentation}
In the previous ablation studies, we used the default data augmentation techniques for training and testing.
For simplicity, we set the original size to $128\times128$ and the crop size to $112\times112$. 
But a larger resolution of original frames can provide more structure information for modeling and more crops for data augmentation.
Therefore, we perform a detailed study on the input frame resolution to investigate its effect on the TAD performance.
To study its effect on detection performance, we adopt the same settings of using \textbf{96 frames at 3FPS} as the input and enable \textbf{SP} and \textbf{TP} in our PlusTAD.

Specifically, we use two choices to expand the frame resolution. One choice is to use the fixed resolution as  ``$171 \times128$'', and the other is to use ``short-128'' to keep the original aspect ratio. 
Among them, ``$171 \times128$'' were first adopted in~\citep{r-c3d} for network training, and we follow its setting to increase the input resolution.
First, we only change the frame resolution and fix the crop size as $112 \times 112$.
As shown in the first three rows in Table~\ref{table:BasicTAD Plus-origin-spatial-size} for both anchor-based and anchor-free PlusTAD, a larger original size can contribute to a better TAD performance. Notably, it can bring around 4\% performance improvement for both anchor-based and anchor-free detection methods. This significant performance improvement indicates that larger input resolution can provide more detailed spatial structure information and generate more diverse training samples.

Furthermore, we perform another comparative study by increasing the crop size. More concrete, we increase the origin frame size to ``short-180'' and crop size to ``$160 \times 160$''. This setting has a similar ratio of crop size and original size to the original PlusTAD. As shown in Table~\ref{table:BasicTAD Plus-origin-spatial-size}, under this setting, the average mAP obtains an improvement of about 2.0\%. This performance improvement demonstrates the effectiveness of increasing input frame resolution and keeping the resolution correspondence between the original frame and crop patch. Therefore, we choose the input frame resolution as ``short-180'' and the crop size as $160 \times 160$ by default in the following studies.
\begin{table*}[ht]
\centering
\small
\caption{\textbf{Ablation Study results on spatial and temporal resolution on THUMOS14~\citep{THUMOS14}.} Overhead represents the increase in computational overhead after expanding the temporal or spatial resolution, and the basic one denotes by $\times1$. }
\label{table:resolution}
\setlength\tabcolsep{1.6mm}
\resizebox{1.0\textwidth}{!}{
\begin{tabular}{l|c|c|cc| cccccc|c}
\toprule
Method &Original Size& Crop Size & FPS&Frames  & mAP@0.3 & mAP@0.4 & mAP@0.5 & mAP@0.6 & mAP@0.7  & $\text{Avg}$ & FLOPs\\
 \midrule
\multirow{4}{*}{Anchor-based}&short-128&$112\times112$   &3& 96  & 68.4&	65.0	&58.6&	49.2&	33.5& 54.9& 136.4G \\
~&short-180&$160\times160$   &3&96&	71.2&	66.8&	61.2&	50.1&	36.3 &57.1 &262.1G\\
~&short-128&$112\times112$    &6& 192   & 70.2&66.3&	60.6&	50.3&	36.2	& 56.7& 272.9G\\
~&short-180&$160\times160$ &6& 192 & \textbf{72.3}  & \textbf{68.4}   &  \textbf{62.0}   &  \textbf{52.4}   &  \textbf{37.0}  &  {\bf 58.4} &519.3G \\

\midrule
\multirow{4}{*}{Anchor-free} &short-128&$112\times112$   &3& 96   &  70.4   &   65.5  &  57.6 & 46.0  & 33.2  &54.5 & 151.5G \\
~&short-180&$160\times160$   &3& \multirow{1}{*}{96}      &  72.5   &  67.2   &  59.1  & 48.2  &  35.3 & 56.4 & 275.9G\\
~&short-128&$112\times112$   &6& \multirow{1}{*}{192}    &  71.7   &  66.9   &  59.0  &  49.2 &  35.3 &56.4  & 284.2G\\
~&short-180&$160\times160$    &6& \multirow{1}{*}{192}   &  \textbf{75.5}   &  \textbf{70.8}   &  \textbf{63.5}   &  \textbf{50.9}   &  \textbf{37.4} & {\bf 59.6} & 533.1G \\
\bottomrule
\end{tabular}
}
\end{table*}

\subsubsection{Effectiveness of Testing Augmentation}
After the ablation on the spatial resolution of training frames and crop patches, we investigate the influence of testing augmentation on the TAD performance. The augmentation methods in the testing phase can improve the robustness of the predictions without additional training costs.
We conduct ablation experiments to study the effect of different cropping methods.
We attempt to apply four following kinds of testing augmentation. 
``CenterCrop'' is cropping the center area the same size as the training crop from images. 
``ThreeCrop'' is cropping three areas of the same size as the training crop from the frames along the long side.
``Flip'' is flipping the frames for horizontal augmentation. 
``Backward'' complements a reverse sliding window process to sample windows. 
Each augmentation method produces a new clip view to be inferred by the network separately and fuses predictions of these views to produce the final TAD results.
Specifically, for ThreeCrop and Flip augmentation, we fuse the features after the neck. For backward augmentation, we fuse the detection results in the post-processing phase.

The results of different testing augmentation are shown in Table~\ref{table:test-augmentation-method}.
We find that ``ThreeCrop'' and ``CenterCrop+Flip'' achieve very similar results for both anchor-based and anchor-free TAD methods, which can improve the performance of ``CenterCrop'' by around 0.5\%.
Unexpectedly, ``+Backward'' seems to hurt the final precision. A possible explanation is that increasing the number of predicted actions degenerates the distribution of the original prediction. Considering the extra computation cost brought by extra testing views and its small performance improvement, we set the default testing scheme of our PlusTAD as the simple ``Center Crop''.

\subsubsection{Study on the spatial and temporal resolutions}
From previous studies, we have concluded that larger frame resolution and more input frames can contribute to a higher TAD performance. We further perform ablation studies to investigate how to keep a balance between spatial resolution and temporal frame under a fixed computational budget. 
Considering video data contains two dimensions, spatial and temporal. Using \textbf{96 frames at 3FPS} as a baseline, we can add sources of information along two alternative paths, one of which is increasing the number of frames (higher temporal resolution) and the other is using larger frames (higher spatial resolution).
We conduct experiments to explore the gain of different data sources and search for an optimal route to boost performance.

As reported in Table~\ref{table:resolution}, doubling the number of frames and doubling the frame size separately brings similar performance improvement for both detection heads. 
It is worth noting that, compared with more frames, a larger resolution of frames may deliver more benefits to the anchor-based method.
When we double the number of frames and double the frame size, both detection heads are further improved and create a new record of temporal action detection. We obtain the mAP of 59.6\% for anchor-free PlusTAD. 
In summary, we find that temporal resolution and spatial resolution are both important for improving TAD performance, and we can choose a reasonable baseline method under the computational resource available.

\begin{table*}[t]
\centering
\small
\caption{\textbf{Summary of configurations} of the optimal PlusTAD on the datasets of THUMOS14, FineAction, and ActivityNet-v1.3. } 
\setlength\tabcolsep{4.1mm}{
\resizebox{1.0\textwidth}{!}{
\begin{tabular}{cccccc}

\toprule 
Dataset & Configuration & Frame Representation & Temporal Resolution & Spatial Resolution & Window Size \\ \midrule
THUMOS14    &    $\text{PlusTAD}_{3,96}^{112}$    &  Window    &   96    &   11$2\times$112     &   32s   \\
THUMOS14    &    $\text{PlusTAD}_{6,192}^{160}$    &  Window    &    192      &   160$\times$160    &     32s        \\
FineAction   &    $\text{PlusTAD}_{2,96}^{224}$    &    Window     &   96     &     224$\times$224    &     48s   \\
ActivityNet-v1.3     &    $\text{PlusTAD}_{96}^{224}$   &   Video    &   96    &    224$\times$224    &      -       \\\bottomrule
\end{tabular}
}
}
\label{table:dataset_config}
\end{table*}

\subsection{Comparison with the State of the Art}
\label{comparisonwiththestateoftheart}

\begin{table*}[t!]
\caption{\textbf{Comparison with state of the art on the THUMOS14.} ``RGB-Only" means whether to use other input modalities besides RGB input. 
} 
\centering
\small

\setlength\tabcolsep{1.8mm}
\resizebox{1.0\textwidth}{!}{
\begin{tabular}{c|c|c |c|c c c c c|c|c}
\toprule
 \multirow{2}*{Type} &  \multirow{2}*{Method} &\multirow{2}*{Backbone} & \multirow{2}*{RGB-Only} & mAP & mAP & mAP & mAP & mAP& mAP &\multirow{2}*{FLOPs} \\
  &  & &  & @0.3 & @0.4 & @0.5 & @0.6 & @0.7& $\text{@Avg}$ & \\
 \midrule
\multirow{11}*{Multi-stage}&BSN~\citep{bsn}&TSN &  \XSolidBrush   &  53.5  &  45.0 &  36.9  & 28.4  &  20.0 &36.8 & -\\
  &MGG~\citep{mgg}&TSN  &  \XSolidBrush   &   53.9  &  46.8 & 37.4 & 29.5 & 21.3&37.8 &-\\
  &BMN~\citep{bmn} &TSN &  \XSolidBrush   &  56.0   &  47.4   &  38.8  & 29.7 &  20.5&38.5 &-\\
  &DBG~\citep{bmn} &TSN &  \XSolidBrush   &  57.8   &  49.4   &  39.8  & 30.2 &  21.7&39.8 &-\\
  &RTD-Net~\citep{rtd} &I3D  & \XSolidBrush   &  58.5   &   53.1  & 45.1 &  36.4 &  25.0&43.6 & -\\
  &TCANet~\citep{tcanet} &TSN &  \XSolidBrush   & 60.6  &  53.2  & 44.6 & 36.8 & 26.7 &44.4&-\\
  &G-TAD~\citep{g-tad} &TSN & \XSolidBrush &  66.4   &  60.4  & 51.6 &  37.6  & 22.9&47.8 &-\\
  &AFSD~\citep{afsd} &I3D &  \XSolidBrush   &  67.3   &  62.4   &    55.5 &  43.7  & 31.1   &  52.0 &2780.0G\\ 
  &DCAN~\citep{DCAN}  & TSN & \XSolidBrush   &   68.2  & 62.7  & 54.1 & 43.9  &  32.6 &52.3 &-\\ 
  & SP-TAD~\citep{sparse-rcnn-tad}&I3D  &  \XSolidBrush   &   69.2  & 63.3  & 55.9 & 45.7  & 33.4  &53.5 &-\\ 
  & TadTR~\citep{e2e-TADTR}&R50-SlowFast &  \CheckmarkBold   &   69.4  & 64.3  & 56.0 & 46.4  & 34.9  &54.2& 475.0G\\ 
  \midrule
  \multirow{12}*{One-stage} & SSAD~\citep{ssad} & TSN & \XSolidBrush & 43.0 & 35.0 & 24.6  & - & - & - & -\\
  &DBS~\citep{dbs} &TSN& \XSolidBrush  & 50.6  & 43.1 & 34.3 & 24.4 & 14.7 &33.4 &-\\
  &A2Net~\citep{a2net} &I3D &  \XSolidBrush  &  58.6   &  54.1  & 45.5 &  32.5  & 17.2&41.6 &-\\
  &PBRNet~\citep{pbrnet} &I3D &  \XSolidBrush  &  58.5   &  54.6  & 51.3 &  41.8  & 29.5&47.1  &-\\
    \cline{2-11} 
  & R-C3D~\citep{r-c3d} & C3D &\CheckmarkBold & 44.8 & 35.6 & 28.9 & - & -  & - &1360.0G\\
  &GTAN~\citep{gtan}&P3D & \CheckmarkBold & 57.8  &  47.2  & 38.8   &-&-&-&-\\
  &DaoTAD~\citep{rgb_enough}&R50-I3D  &  \CheckmarkBold   &  62.8   &  59.5   &  53.8   &  43.6  &  30.1&50.0&206.7G\\ 
  &DaoTAD$_{3,96}^{112}$~\citep{rgb_enough}&R50-SlowOnly  &  \CheckmarkBold   &  63.2   &  59.7   &  54.4   &  45.6  &  32.2 &51.0& 133.3G\\
  & \textbf{$\text{PlusTAD}_{3,96}^{112}$(Anchor-based)}&R50-SlowOnly & \CheckmarkBold & \textbf{68.4}  &  \textbf{65.0} &  \textbf{58.6}   &  \textbf{49.2}   & \textbf{33.5} &  \textbf{54.9} &136.4G\\

  &\textbf{ $\text{PlusTAD}_{3,96}^{112}$(Anchor-free)}&R50-SlowOnly & \CheckmarkBold & \textbf{70.4}    & \textbf{65.5}    & \textbf{57.6}    & \textbf{46.0}    & \textbf{33.2}    & \textbf{54.5} &151.5G\\ 
  & \textbf{$\text{PlusTAD}_{6,192}^{160}$(Anchor-based)}&R50-SlowOnly & \CheckmarkBold & \textbf{72.3}  &  \textbf{68.4} &  \textbf{62.0}   &  \textbf{52.4}   & \textbf{37.0} &  \textbf{58.4} &519.3G\\

  &\textbf{$\text{PlusTAD}_{6,192}^{160}$(Anchor-free)}&R50-SlowOnly & \CheckmarkBold & \textbf{75.5}    & \textbf{70.8}    & \textbf{63.5}    & \textbf{50.9}    & \textbf{37.4}    & \textbf{59.6} &533.1G\\ 
\bottomrule
\end{tabular}
}
\label{table:sota_thumos14}
\end{table*}
In previous subsections, we have performed thorough ablation studies on the basic design in our modular BasicTAD framework. We also provide extensive investigation on our proposed new design principles to enhance the TAD performance of BasicTAD. In this subsection, we turn to compare our PlusTAD with previous state-of-the-art methods. In this comparison, we direct transfer these optimal configurations discovered from THUMOS14~\citep{THUMOS14} to the other two large-scale benchmarks: FineAction~\citep{fineaction} and ActivityNet v1.3~\citep{anet}. 
Table~\ref{table:dataset_config} reports our configurations on the three datasets. 
We use $\text{PlusTAD}_{\text{R},\text{F}}^{\text{S}}$ to represent the configuration name on the datasets with the window sampling strategy, such as THUMOS14 and FineAction.
$\text{R},\text{F},\text{S}$ is the frame rate of the sampling, the number of frames, and the size of the frames.
For the datasets based on the video-level sampling strategy, such as ActivityNet v1.3, we delete $\text{R}$ of the subscript due to the sampled frames representing the whole video. We compare these configurations with state-of-the-art methods on the three datasets. 

\textbf{THUMOS14.}
We compare other state-of-the-art methods in Table~\ref{table:sota_thumos14} on the THUMOS14 dataset. 
We compute the FLOPs of other end-to-end methods~\citep{afsd,rgb_enough,e2e-TADTR,r-c3d} in the table for a fair comparison with our method. 
For some two-stage and head-only methods, their FLOPs are unavailable and indicated as ``-'' in the table.
From this table, we see that our PlusTAD significantly outperforms previous methods by a large margin with only RGB as input. 
In particular, $\text{PlusTAD}_{6,192}^{160}$ uses more frames with a larger size and improves the mAP jumps by 4 to 5 points.
Meanwhile, our overall FLOPS are still less than these methods, and our simple design allows for very fast deployment.
These superior results demonstrate the effectiveness of our PlusTAD thanks to its simplicity and end-to-end training.

\begin{table*}[t]
\centering
\small
\caption{\textbf{Comparison with state of the art on the FineAction dataset.} ``RGB-Only" means whether to use other input modalities besides RGB input.}
\setlength\tabcolsep{4.45mm}
\resizebox{1.0\textwidth}{!}{
\begin{tabular}{c|c|c|ccc|c}
\toprule
Method &Backbone&RGB-Only& mAP@0.5 & mAP@0.75 & mAP@0.95 & Avg\\
\midrule
BMN~\citep{bmn} &I3D &\CheckmarkBold&  12.56   & 7.49 &   2.62  & 7.86 \\
BMN~\citep{bmn} &I3D &\XSolidBrush &  14.44   & 8.92 &   3.12  & 9.25 \\
DBG~\citep{dbg} &I3D &\CheckmarkBold&  8.57   & 5.01 &    1.93  & 5.31 \\
DBG~\citep{dbg} &I3D &\XSolidBrush &  10.65   &  6.43 &   2.50  &  6.75 \\
G-TAD~\citep{g-tad}&I3D  &\CheckmarkBold&  10.88   &  6.52 &    2.19  & 6.87 \\
G-TAD~\citep{g-tad}&I3D  &\XSolidBrush &  13.74   &   8.83 &    \textbf{3.06} &  9.06 \\
\midrule
\textbf{$\text{PlusTAD}_{2,96}^{224}$(Anchor-based)} &R50-SlowOnly&\CheckmarkBold     &  \textbf{24.34}   & \textbf{10.57} &   0.43  & \textbf{12.15} \\
\textbf{$\text{PlusTAD}_{2,96}^{224}$(Anchor-free)} &R50-SlowOnly&\CheckmarkBold     &  22.37  & 10.36 &  0.83   & 11.66 \\
\bottomrule
\end{tabular}
}
\label{table:fineaction}
\end{table*}

\begin{table*}[t]
\centering
\small
\caption{\textbf{Comparison with state of the art on the ActivityNet-1.3 dataset.} ``RGB-Only" means whether to use other input modalities besides RGB input. }
\setlength\tabcolsep{4.45mm}
\resizebox{1.0\textwidth}{!}{
\begin{tabular}{c|c|c|c c c | c}
\toprule
Method &Backbone&RGB-Only& mAP@0.5 & mAP@0.75 & mAP@0.95 & Avg\\
\midrule
 BSN~\citep{bsn} &TSN &\XSolidBrush&  46.45   & 29.96 &    8.02  & 30.03 \\
 P-GCN~\citep{pgcn}&-  &\XSolidBrush&  48.26   &  33.16 &    3.27  & 31.11 \\
 BMN~\citep{bmn} &TSN &\XSolidBrush &  50.07   & 34.78 &   8.29  & 33.85 \\
 
 G-TAD~\citep{g-tad}&TSN  &\XSolidBrush&  50.36   &  34.60 &    \textbf{9.02}  & 34.09 \\
 AFSD~\citep{afsd} &I3D &\XSolidBrush &\textbf{52.40} & \textbf{35.30}& 6.50& \textbf{34.40}\\
 SP-TAD~\citep{sparse-rcnn-tad} &I3D &\XSolidBrush &50.06&32.92&8.44 &32.99 \\
\midrule

 BMN~\citep{bmn} &TSN &\CheckmarkBold &  41.93   & 30.10 &   \textbf{9.00} & 29.23 \\
 G-TAD~\citep{g-tad}& TSN &\CheckmarkBold&  45.68   &  31.36 &   7.42  & 30.98 \\
 AFSD~\citep{afsd} &I3D &\CheckmarkBold &- & -& -& 32.90\\ 
\textbf{ $\text{PlusTAD}_{96}^{224}$(Anchor-based)}&R50-SlowOnly &\CheckmarkBold     &  50.04   & \textbf{33.79} & 2.75    & 32.13 \\
 
\textbf{ $\text{PlusTAD}_{96}^{224}$(Anchor-free)}&R50-SlowOnly &\CheckmarkBold     &  \textbf{51.20}  & 33.41 &  7.57   &  \textbf{33.12} \\

\bottomrule
\end{tabular}
}
\label{table:anet}
\end{table*}

\textbf{FineAction.}
We also compare our proposed PlusTAD with the state-of-the-art methods on FineAction~\citep{fineaction}. FineAction is a new and large-scale TAD dataset, and few works have reported results on this new benchmark.
As shown in Table~\ref{table:fineaction}. 
We compare our $\text{PlusTAD}_{2,96}^{224}$ with three representative works~\citep{bmn,dbg,g-tad}, which are provided by~\citep{fineaction}. All of them are based on pre-extracted I3D features. We find that our method can obtain a better performance on Average mAP. 
However, our method performs worse on mAP@0.95. 
One possible reason is that all three methods above are multi-stage methods that can get refined proposals from the previous detection or proposal generation stage. This leads to better localization results for metrics at high thresholds like mAP@0.95.

\textbf{ActivityNet-v1.3.}
Due to the complex semantics of the action class on ActivityNet-v1.3, we adapt our PlusTAD to generate binary action proposals and obtain the detection results by applying video-level action classifiers. This is a common setting for many previous state-of-the-art methods on ActivityNet-v1.3.
This is mainly due to the annotation sparsity of ActivityNet-v1.3, and thus the detection results can benefit from the video-level classification.
As shown in Table~\ref{table:dataset_config}, due to the sparse action instance distribution of ActivityNet-v1.3, we sparsely sample 96 frames from the entire video to represent the whole video. This sparse sampling would greatly increase the temporal receptive field but also lose detailed temporal information that might be useful for TAD.

As shown in Table~\ref{table:anet}, our $\text{PlusTAD}_{96}^{224}$ performs better than~\citep{bmn,g-tad,afsd} when only using RGB frames as input.
Meanwhile, our method is a one-stage detector without any refinement sub-network, such as AFSD~\citep{afsd}, to refine boundaries. 
However, other head-only methods use pre-extracted features from the entire video to get richer information. Note in this case, they can use more frames (not only 96) for feature extraction.
Their detection results can be further improved by introducing an additional optical flow modality.
Considering these several factors, our PlusTAD still remains a powerful baseline method due to its simplicity, effectiveness, and efficiency.

\subsection{Efficiency Analysis}
\label{efficiencyanalysis}
\begin{table*}[t]
\centering
\small
\caption{\textbf{Comparison of inference speed.} The running speed of previous methods is directly cited from their papers. Note that these methods all use optical flow as inputs, and their running time does not include the optical flow calculation. So, their real running speed is lower than the reported speed. $\dagger$ donate DaoTAD~\citep{rgb_enough} tested in our platform.
}
\setlength\tabcolsep{4.4mm}
\resizebox{1.0\textwidth}{!}{
\begin{tabular}{c|ccc|c|c|c|c}
\toprule
Method &FPS&Frames&Size& GPU      & RGB-Only& FPS & mAP \\ \midrule
SS-TAD~\citep{end2end}&-&-&- & TITAN XM &\XSolidBrush&  $<$ 701 & - \\
R-C3D~\citep{r-c3d} &25&768&$112\times112$ & TITAN XP &\XSolidBrush& $<$ 1030& - \\
PBRNet~\citep{pbrnet}&10 &256 &$96\times96$ & 1080Ti   &\XSolidBrush& $<$ 1488 & 47.1\\
AFSD~\citep{afsd} &10&256&$96\times96$   & 1080Ti   &\XSolidBrush& $<$ 3259& 52.0 \\
AFSD~\citep{afsd}  &10&256&$96\times96$ & V100    &\XSolidBrush& $<$ 4057 & 52.0\\
DaoTAD~\citep{rgb_enough} &25&768&$112\times112$ & 1080Ti &\CheckmarkBold &6668&50.0\\
SP-TAD~\citep{sparse-rcnn-tad} &10&256&$96\times96$ & V100 &\XSolidBrush&$<$5574&53.5\\
TadTR~\citep{e2e-TADTR} &10&256&$96\times96$&TITAN XP&\CheckmarkBold& 5076&54.2\\
\midrule
DaoTAD$\dagger$~\citep{rgb_enough} &25&768&$112\times112$ & TITAN XP &\CheckmarkBold &7064&50.0\\
DaoTAD$\dagger$~\citep{rgb_enough} &25&768&$112\times112$ & V100 &\CheckmarkBold &8989&50.0\\
\textbf{$\text{PlusTAD}_{3,96}^{112}$(Anchor-based)}  &3&96&$112\times112$    & TITAN XP  &\CheckmarkBold &  \textbf{13715} &  \textbf{54.9} \\ 
\textbf{$\text{PlusTAD}_{3,96}^{112}$(Anchor-based)} &3&96&$112\times112$  &    V100    & \CheckmarkBold&    \textbf{17454}  & \textbf{54.9}\\
\textbf{$\text{PlusTAD}_{3,96}^{112}$(Anchor-free)}  &3&96& $112\times112$    & TITAN XP &\CheckmarkBold &  \textbf{7143} &  \textbf{54.5} \\ 
\textbf{$\text{PlusTAD}_{3,96}^{112}$(Anchor-free)}  &3&96&$112\times112$  &    V100    & \CheckmarkBold&    \textbf{8377}  & \textbf{54.5}\\
\bottomrule
\end{tabular}
}
\label{table:inferencespeed}
\end{table*}

In the above subsections, we have demonstrated that our BasicTAD and PlusTAD can achieve very high TAD mAP on the standard benchmarks. Due to its simplicity in design, our PlusTAD enjoys high efficiency as well. In this subsection, we report the running speed of our PlusTAD and compare it with the other state-of-the-art methods with end-to-end training strategies.

To make the comparison fair and rigorous, we first define the inference speed for TAD tasks. 
We use ``FPS'' to represent the number of frames per second processed by the model when processing the video stream. 
``FPS'' here is calculated differently between the image field and TAD. Given an input video with $s$ seconds and its $f$ FPS, the total number of frames to be processed is $s\times f$. If we re-sample frames before feeding them into the network, we can get a new input with $s$ seconds, $f'$ FPS, and $s \times f'$ frames. Suppose we spend $t$ seconds to process re-sampled frames. The speed (FPS) of processing the origin video is $\frac{s \times f}{t}$.

We use 96 frames at 3FPS to evaluate the inference speed of PlusTAD with both heads because the average mAP is close to other SOTA methods under this setting. Considering using the same GPU, as shown in Table~\ref{table:inferencespeed}, PlusTAD with the anchor-based method yields 17454 FPS using V100 GPU, which is $4 \times$ faster than~\citep{afsd}. 
Although PlusTAD with the anchor-free method is much slower because of more channels in the neck and head, it still has a higher FPS and mAP than AFSD~\citep{afsd} and TadTR~\citep{e2e-TADTR}. 

Furthermore, to eliminate the influence of other hardware, such as CPU, RAM, and disk, we also compare the inference speed of DaoTAD~\citep{rgb_enough} reproduced on our platform. In Table~\ref{table:inferencespeed}, the FPS on 1080Ti reported by DaoTAD~\citep{rgb_enough} is slower by 5.9\% than our report. It roughly matches the performance gap between 1080Ti and TITAN XP. From the above analysis, our PlusTAD still maintains an advantage in inference speed.

These comparisons show that it is necessary to re-design a simple and efficient TAD baseline method, and our proposed PlusTAD can achieve a new astounding baseline of high effectiveness and efficiency.

\begin{figure*}[!t]
  \includegraphics[width=1\textwidth]{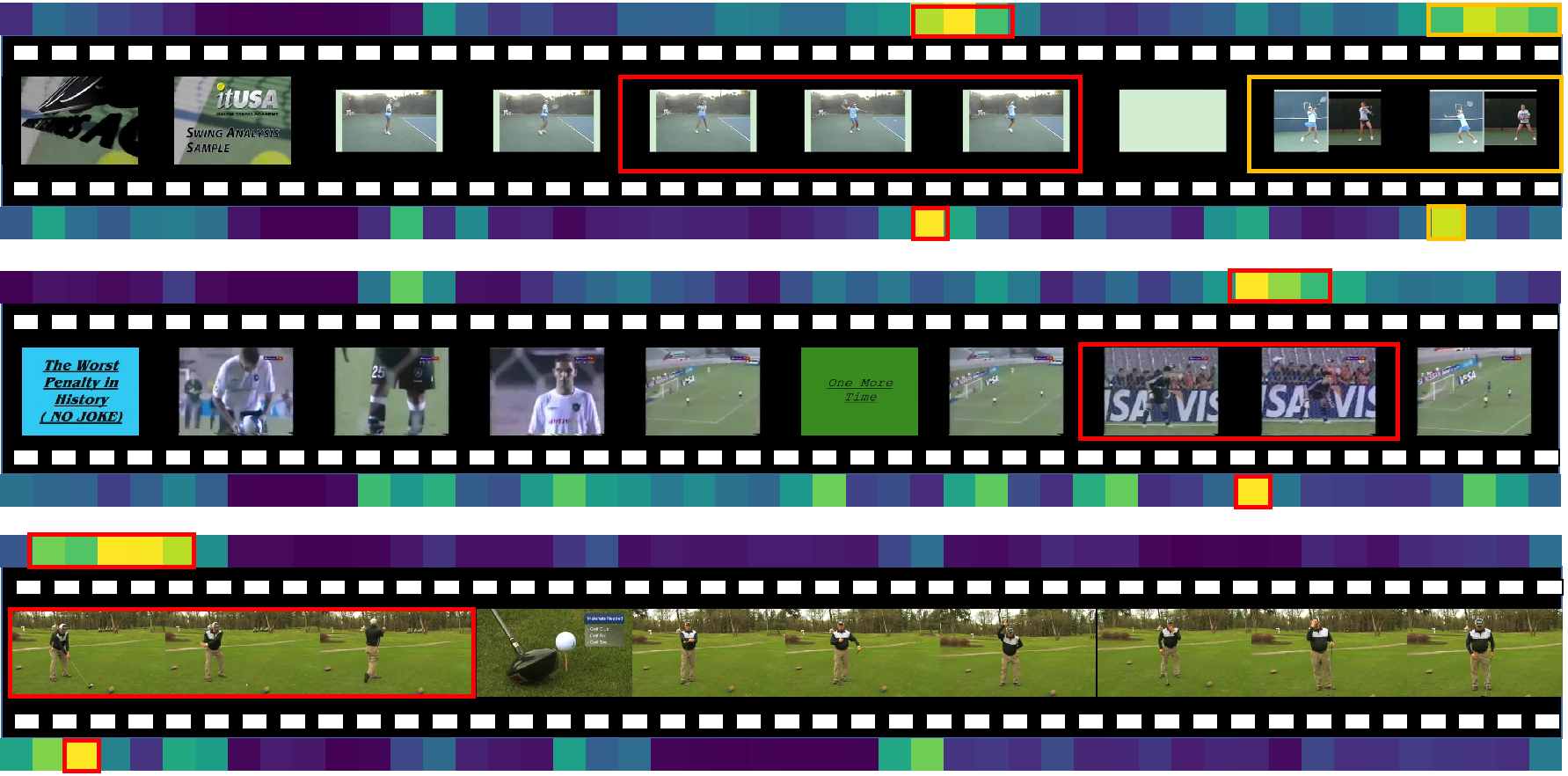}
  \caption{Visualization of our anchor-based and anchor-free methods on three different actions ``TennisSwing'', ``SoccerPenalty'' and ``GolfSwing''. For each action, the top feature map represents the anchor-based method's action activation feature map, while the bottom represents the anchor-free method's action activation map. The middle one is the original action frame, and the video order is from left to right. For example, we choose ``TennisSwing'' with two ground truth actions in the red and orange boxes. Anchor-based and anchor-free methods each have two activation parts of target actions, and we mark them with red and orange boxes.
}
  \label{fig:crop}
\end{figure*}

\subsection{Feature Activation Analysis}
\label{featureactivationanalysis}
After conducting extensive quantitative studies on the accuracy and efficiency of our proposed BasicTAD and PlusTAD, we present some visualization analysis on THUMOS14 in this subsection. We start by visualizing the feature maps of the backbone for both anchor-based and anchor-free heads. To enhance the visualization clarity, we perform average pooling to squeeze the feature map spatially and retain only the temporal dimension. As illustrated in Figure~\ref{fig:crop}, we depict the feature activation of anchor-based PlusTAD at the top, the original frames with ground truth in the middle, and the feature activation of anchor-free PlusTAD at the bottom. We provide three examples in total.
As evident in Figure~\ref{fig:crop}, the anchor-based method exhibits a preference for activating a large area of target regions. On the other hand, the anchor-free method demonstrates activation at a specific and single location within the target action.
Figure~\ref{fig:crop} shows that the anchor-based method prefers to activate at a large area of target regions. 
In contrast, the anchor-free method activates at a specific and single location inside the target action. While these findings cannot be considered definitive conclusions, we believe they can serve as useful guidelines for future research to investigate the application of anchor-based and anchor-free methods in TAD and draw more comprehensive conclusions.

\begin{figure*}[htbp!]  
	\centering
	\subfigure{
		\includegraphics[width=0.35\linewidth]{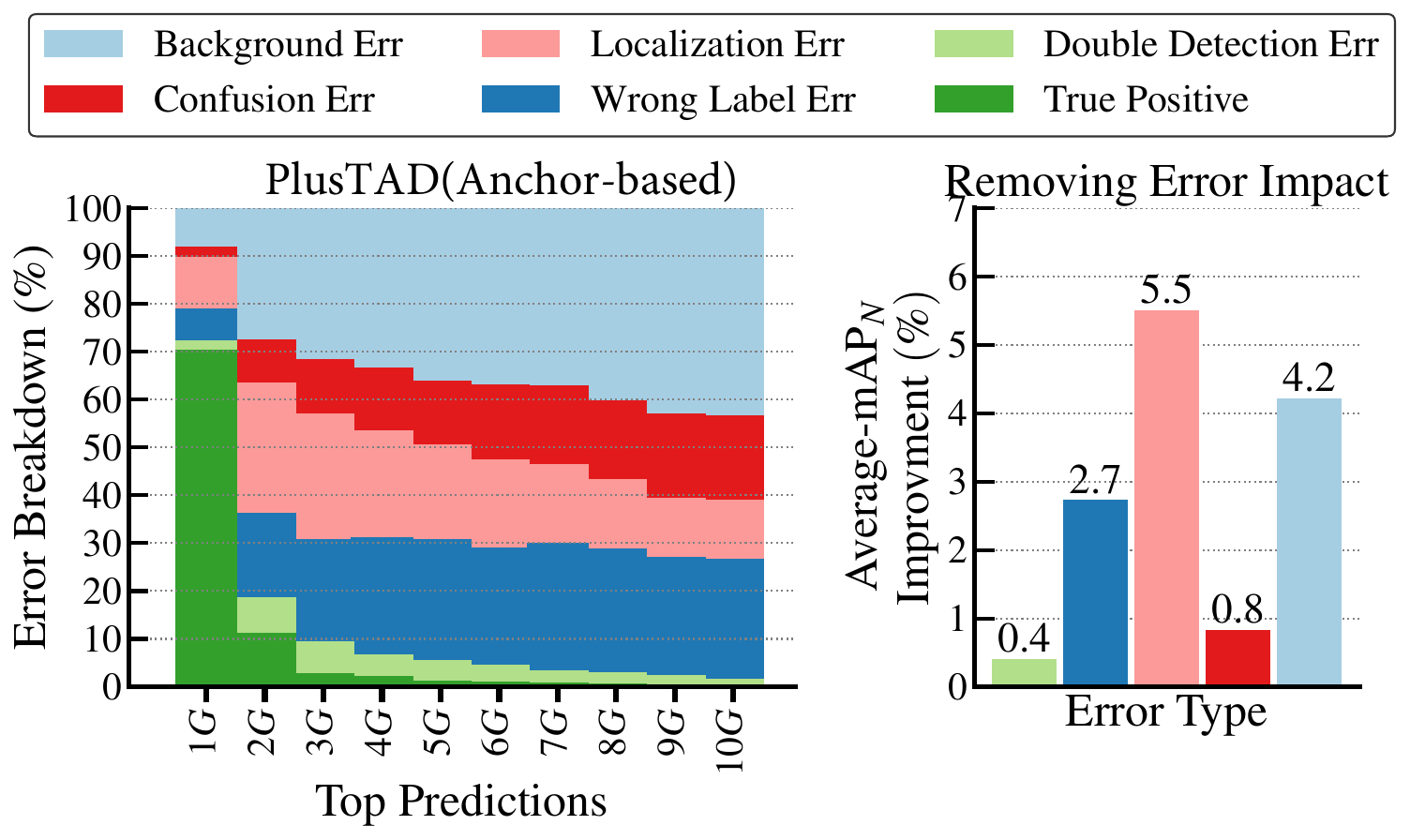}
	}
	\subfigure{
		\includegraphics[width=0.56\linewidth]{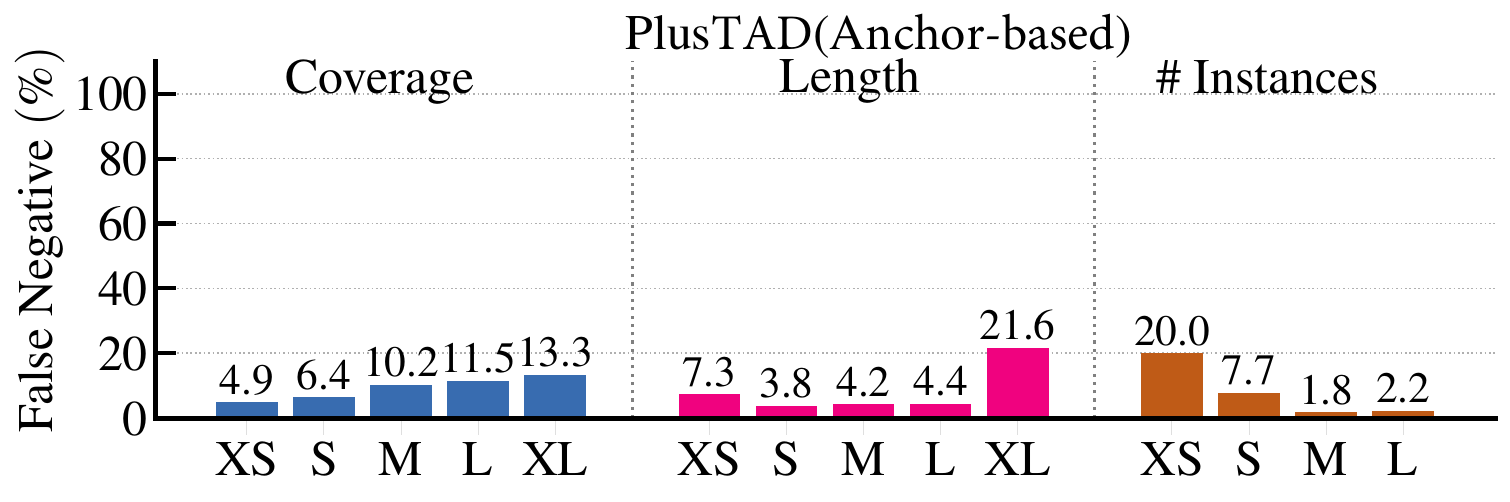}
	}
    
    \vspace{-4mm}
    \setcounter{subfigure}{0}
	\subfigure[False Positive Profiles]{  
		\includegraphics[width=0.35\linewidth]{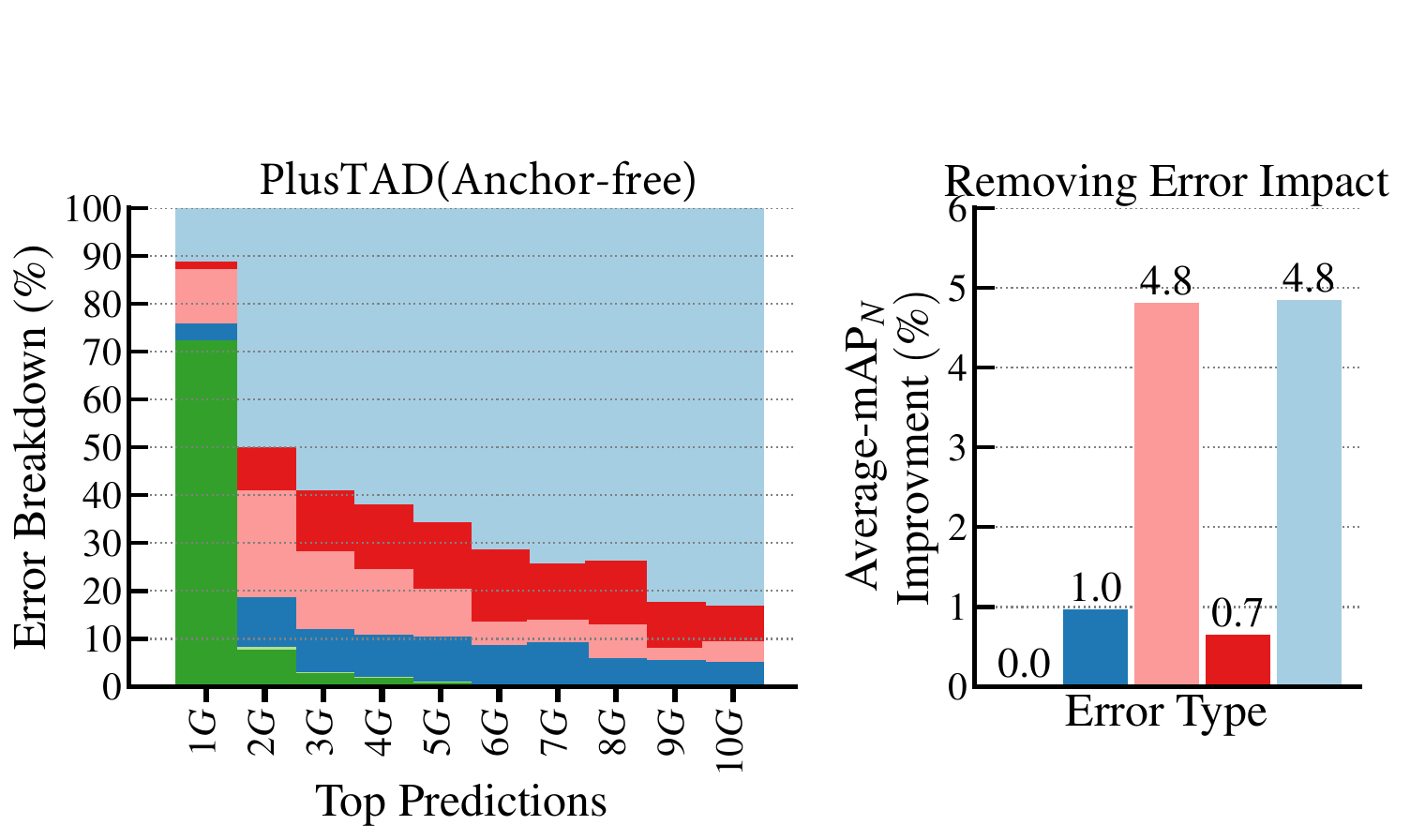}
	}
	\subfigure[False Negative Profiles]{  
		\includegraphics[width=0.56\linewidth]{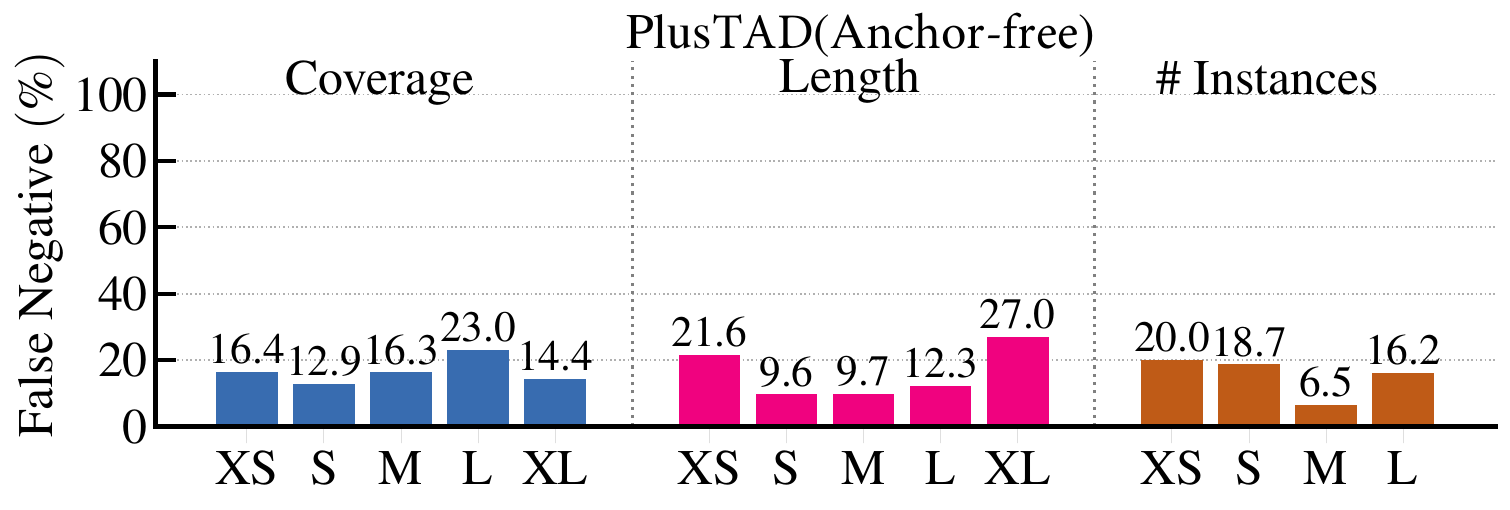}
	}
	\caption{(a) \textbf{False Positive Profiles.} Left: False positive profiles of both methods. Each profile demonstrates the FP error breakdown in the top-10G predictions. Right: Improvement gained from removing all predictions that cause each type of error. The higher the value, the greater the effect on average-mAP$_{N}$.
  (b) \textbf{False Negative Profiles.} Average false negative rate across algorithms for each characteristic on both methods. }
  \label{fig:fp_fn}
	
\end{figure*}

\subsection{Error Analysis}
In addition to the feature map visualization, we provide the error analysis on our detection results as well.
To better understand the detection errors, we use the diagnostic tool provided by~\cite{detad} to analyze the detection results of PlusTAD with 192 frames at 6FPS. 
We visualize the error analysis of both anchor-based and anchor-free methods to make a direct comparison between them and provide some insightful analysis of both TAD methods.
\paragraph{False Positive Analysis}
The left of Figure~\ref{fig:fp_fn} shows false positive (FP) profiling. 
As shown on the left of the figure, top-1G predictions contain the most TP (true positive)  predictions. The majority of error type is localization error and background error. When we consider more predictions, the true positive rate will decrease dramatically. In this sense,  the vast majority of ground truth actions can be successfully predicted within top-1G predictions. 
Comparing the anchor-based and anchor-free methods from top-2G to top-10G, we could see that the anchor-free method suffers more from ``Background Err''.
It may be due to limited anchors causing more predictions to fail to match ground truth.  
Conversely, the anchor-based method has sufficient anchors to reduce ``Background Err'', but may bring other errors such as ``Wrong Label Err''.

\paragraph{False Negative Analysis}
False negative (FN) profiling is illustrated in the right of Figure~\ref{fig:fp_fn}.
From the visualization, we see that under the measure of Coverage and Length, the FN rate of the anchor-based method is significantly lower than the anchor-free method. Since short actions of the same category usually have more numbers in THUMOS14, the anchor-based method can benefit from the dense sample matching mechanism and tends to better detect short actions. While, for the anchor-free method, it lacks the flexibility of detecting very small or large action instances, which might be due to the difficulty of directly regressing the boundaries of these hard action instances. In general, we can conclude that the anchor-based method might be more robust than the anchor-free method for dealing with extremely hard cases, but not their detection results are not as accurate as the anchor-free method.

\section{Conclusion and Future Work}
\label{conclusions}
In this paper, we have reconsidered the design of the TAD pipeline and presented a simple modular detection framework. Based on this modular design, we perform extensive investigation over the basic options in each component and finally, come up with a simple yet effective TAD baseline, termed BasicTAD. Furthermore, we improve the BasicTAD with minimal changes by following the core design of preserving rich information in the backbone and neck, and the resulted detector is termed PlusTAD. Our end-to-end BasicTAD and PlusTAD are free of pre-processing and can be used in real-time application scenarios. Extensive experiments demonstrate that our PlusTAD significantly outperforms previous state-of-the-art methods on the THUMOS14 and FineAction, and achieves quite competitive results on the ActivityNet-v1.3 datasets. We also provide in-depth ablation studies on each TAD component in a step-by-step manner and detailed visualization results to figure out the main property and major error of our PlusTAD. We hope our approach can serve as a strong baseline for future TAD research.

In the future, we can enhance the temporal modeling capacity of our BasicTAD and PlusTAD by incorporating long-term memory. This addition will enable us to tackle action instances of longer duration with greater precision and accuracy. In addition, we can expand our TAD framework to include Transformer backbones with global receptive fields to boost its representation power. Doing so can significantly improve the final TAD performance, further enhancing our ability to analyze complex temporal sequences.











\bibliographystyle{cas-model2-names}

\bibliography{cas-refs} 



\end{document}